\documentclass[11pt]{article}

\usepackage[preprint]{acl}

\usepackage{times}
\usepackage{latexsym}

\usepackage[T1]{fontenc}

\usepackage[utf8]{inputenc}

\usepackage{microtype}

\usepackage{inconsolata}

\usepackage{graphicx}

\title{PRAGMA: Evaluating Personalized Guidance with Memory Alignment in Lifelong Conversations}
\author{
 \textbf{Hyojeong Yu\textsuperscript{1}},
 \textbf{Hyukhun Koh\textsuperscript{2}},
 \textbf{Minsung Kim\textsuperscript{1}},
 \textbf{Yunah Jang\textsuperscript{1}},
 \textbf{Kyomin Jung\textsuperscript{1,2,$\dagger$}}
\\
\\
 \textsuperscript{1}Dept. of ECE, Seoul National University,
 \textsuperscript{2}IPAI, Seoul National University
\\
 \texttt{\{hyoj.yu, hyukhunkoh-ai, kms0805, vn2209, kjung\}@snu.ac.kr}
}

\usepackage{booktabs}  
\usepackage[table]{xcolor}
\usepackage{tabularx}
\usepackage{makecell}
\usepackage{graphicx}
\usepackage{float}
\usepackage{placeins}
\usepackage{multirow}
\usepackage{tcolorbox}
\tcbuselibrary{breakable, skins}
\usepackage{enumitem}
\usepackage{caption}
\newcommand{\promptgroup}[2]{%
\begin{tcolorbox}[
  enhanced,
  title={\footnotesize\sffamily\bfseries #1},
  colback=gray!4, colframe=gray!50,
  colbacktitle=gray!20, coltitle=black,
  boxrule=0.5pt,
  fontupper=\small\ttfamily,
  before skip=4pt,
  after skip=4pt
]
#2
\end{tcolorbox}}

\usepackage{caption}
\usepackage{needspace}

\begin{document}
\maketitle

\begingroup
\renewcommand\thefootnote{$\dagger$}
\footnotetext{Corresponding author.}
\endgroup

\begingroup
\renewcommand\thefootnote{*}
\footnotetext{Code and data are available at
\url{https://github.com/yuhyojeong/PRAGMA} and
\url{https://huggingface.co/datasets/stellahj/PRAGMA}.}
\endgroup

\begin{abstract}

Large language models (LLMs) are increasingly deployed as personalized assistants that interact with users over extended periods of time.
As conversations grow longer, relying on full interaction histories becomes increasingly inefficient and unreliable: long contexts introduce substantial computational overhead, making it difficult for models to consistently identify and utilize the most relevant information for the current request.
These challenges have motivated memory systems that structure and retrieve user-specific information.
In realistic interactions, users often seek practical guidance such as recommendations, planning, and decision support.
Unlike factual recall tasks, personalized guidance requires models to integrate information across multiple past conversations and reason about changing user preferences and experiences.
However, existing conversational memory evaluations mainly focus on retrieval and factual recall.
To study this challenge, we introduce {\large\textsc{pragma}}, a benchmark for evaluating personalized guidance in long-term conversations.
{\large\textsc{pragma}} contains curated longitudinal conversation histories, evidence annotations, and guidance scenarios grounded in evolving user contexts and incorrect user assumptions.
Experiments across retrieval systems, memory systems, and long-context models reveal that current systems struggle both to recover the appropriate conversational evidence and to effectively use it for personalized guidance.
Our results highlight the need for memory architectures that support robust conversational retrieval and memory-grounded reasoning beyond evidence recall.
\end{abstract}

\section{Introduction}

\begin{figure}[t]
    \centering
    \includegraphics[width=\linewidth]{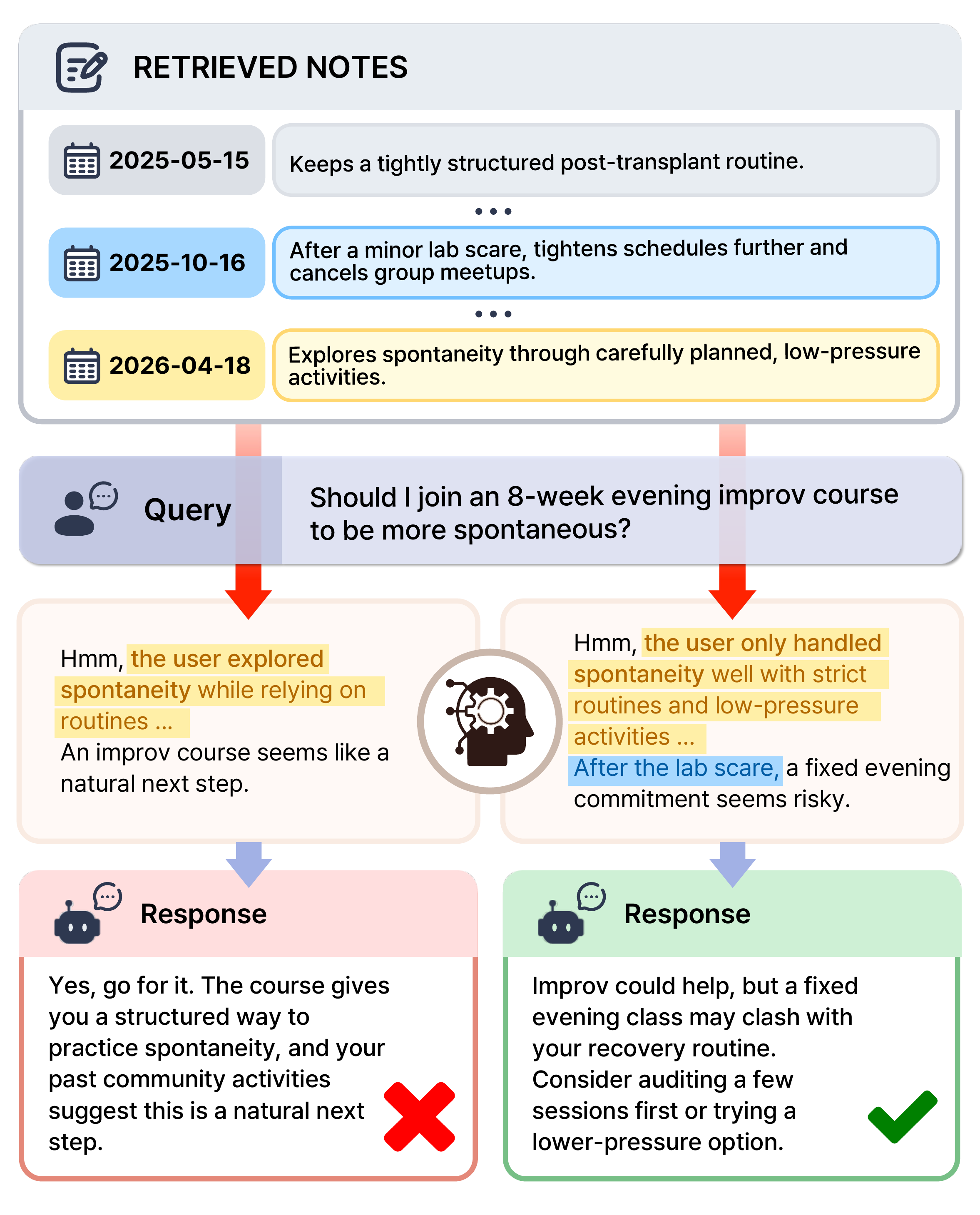}
    \caption{Personalized guidance requires both effective conversational retrieval and downstream memory-grounded reasoning. Relevant evidence may be temporally distributed and only implicitly connected to the final user request. Although both responses use retrieved memories, only the right response correctly synthesizes the user's longitudinal context.}
\vspace{-0.3cm}
    \label{fig:intro}
\end{figure}

\begin{table*}[t]
\centering
\footnotesize
\setlength{\tabcolsep}{3pt}
\renewcommand{\arraystretch}{1.0}

\resizebox{0.72\linewidth}{!}{
\begin{tabular}{lcccccccc}
\toprule
\textbf{Benchmark}
& \textbf{In-situ}
& \textbf{Open}
& \textbf{Guide}
& \textbf{E-A}
& \textbf{E-C}
& \textbf{T-A}
& \textbf{T-C}
& \textbf{Tokens} \\
\midrule

LongMemEval
& $\circ$ & $\scriptstyle\triangle$ & $\scriptstyle\triangle$
& $\circ$ & $\times$ & $\times$ & $\times$
& 115K, 1.5M \\

LoCoMo
& $\times$ & $\times$ & $\times$
& -- & -- & -- & --
& 9K \\

HiCUPID
& $\circ$ & $\circ$ & $\scriptstyle\triangle$
& $\circ$ & $\times$ & $\times$ & $\times$
& 17K \\

ImplexConv
& $\circ$ & $\circ$ & $\circ$
& $\circ$ & $\times$ & $\times$ & $\times$
& 60K \\

ConvoMem
& $\circ$ & $\scriptstyle\triangle$ & $\scriptstyle\triangle$
& $\circ$ & $\times$ & $\times$ & $\times$
& 1K--3M \\

PersonaMem
& $\circ$ & $\times$ & $\circ$
& $\circ$ & $\times$ & $\circ$ & $\times$
& 32K--1M \\

\rowcolor[HTML]{EAF4FB}
\textbf{PRAGMA}
& $\circ$ & $\circ$ & $\circ$
& $\circ$ & $\circ$ & $\circ$ & $\circ$
& 160K \\

\bottomrule
\end{tabular}
}

\caption{
Comparison of long-term conversational memory benchmarks.
In-situ: queries embedded in conversations;
Open: open-ended generation;
Guide: personalized guidance;
E-A/E-C and T-A/T-C: event- and trajectory-grounded aligned/corrective reasoning.
Tokens: approximate context length.
$\circ$: supported, $\triangle$: partial, $\times$: unsupported.
}

\label{tab:benchmark_comparison}
\vspace{-10pt}
\end{table*}

Large language models (LLMs) are increasingly deployed as personalized conversational assistants that interact with users over extended periods of time~\cite{DBLP:conf/naacl/LiYZDWC25, DBLP:conf/acl/0001YHH0LSCPLI025}.
Applications such as recommendation agents, tutors, and productivity assistants rely on awareness of users' preferences, experiences, and evolving needs across interactions.
As conversational histories grow longer, conditioning on full history becomes increasingly inefficient and unreliable, motivating memory systems that selectively store and retrieve user-specific information~\cite{DBLP:conf/aaai/ZhongGGYW24}.

Existing work on conversational memory has primarily focused on retrieval and factual recall from long histories~\cite{DBLP:conf/iclr/WuWYZCY25, DBLP:conf/acl/MaharanaLTBBF24}.
However, real-world personalization often requires practical guidance~\cite{chatterji2025howpeopleuse} such as recommendations, planning support, and decision-making grounded in evolving user experiences.
These queries often require reasoning over long-term user trajectories and potentially incorrect user assumptions~\cite{feng2026doespersonalizedmemoryshape, DBLP:conf/iclr/SharmaTKDABDHJK24}.
Moreover, relevant evidence is often temporally distributed and only implicitly connected to the final user request, making conversational retrieval itself challenging.
As a result, personalized guidance depends not only on retrieving relevant memories, but also on utilizing them coherently during response generation~\cite{kwon2026embodied}.
Yet the relationship between retrieval and downstream personalized reasoning remains underexplored~\cite{laban2026llms, li2026inverseknowledgesearchverifiable}.

Constructing realistic evaluation settings for this problem is also challenging.
Without careful design, synthetic long-term conversations can produce shallow trajectories or queries solvable through simple recency heuristics.
Effective evaluation therefore requires balancing realism, controllability, and resistance to retrieval shortcuts.

To address these challenges, we introduce \textbf{{\large\textsc{pragma}}} (\textbf{PRA}ctical \textbf{G}uidance with \textbf{M}emory \textbf{A}lignment), a benchmark for evaluating personalized guidance in long-term conversations.
{\large\textsc{pragma}} is built through a controlled, human-validated pipeline that generates long-term conversational histories with evolving user states and diverse memory requirements.
The benchmark includes guidance scenarios grounded in both event-level memories and user states that evolve over time, including settings where users make assumptions that conflict with their conversational history.

We evaluate retrieval systems, structured memory systems, and long-context models on {\large\textsc{pragma}}.
Across architectures and generation models, we find that current systems struggle both to recover the appropriate conversational evidence and to effectively use it for personalized guidance.
Even when relevant evidence is retrieved, models often fail to generate coherent and well-grounded responses, particularly on trajectory-grounded and corrective reasoning tasks.
Our findings also highlight the need for memory architectures that support both robust conversational retrieval and downstream memory-grounded reasoning for long-term personalized assistance.

In summary, our contributions are as follows:

\begin{itemize}
    \item We introduce {\large\textsc{pragma}}, a benchmark for evaluating personalized guidance grounded in long-term conversational memory, focusing on guidance scenarios that require reasoning over evolving user trajectories and potentially incorrect user assumptions.
    
    \item We propose a controlled, human-validated benchmark construction pipeline that generates realistic longitudinal conversations with evolving memory dependencies and fine-grained evidence annotations.
    
    \item We evaluate retrieval systems, memory systems, and long-context models on {\large\textsc{pragma}}, showing that current systems struggle to recover relevant memory and to effectively utilize it for personalized guidance.
\end{itemize}

\section{Related Work}

\textbf{Factual memory recall benchmarks.}
Recent benchmarks such as LongMemEval~\cite{DBLP:conf/iclr/WuWYZCY25}, LoCoMo~\cite{DBLP:conf/acl/MaharanaLTBBF24}, and ConvoMem~\cite{pakhomov2025convomembenchmark150conversations} evaluate whether language models can retain and access information from long conversational histories through tasks including factual QA, dialogue understanding, and memory-grounded response generation.
While these benchmarks have advanced evaluation of long-context memory and conversational consistency, they primarily focus on recovering or reproducing past information rather than utilizing memory for open-ended practical guidance.

\textbf{Personalized conversational benchmarks.}
Another line of work studies personalized generation conditioned on user history.
ImplexConv~\cite{DBLP:conf/emnlp/LiBDZS25} evaluates implicit reasoning over semantically distant conversational evidence, but focuses on narrow reasoning settings.
PersonaMem~\cite{Jiang2025KnowMR} introduces preference evolution and recommendation scenarios, yet relies on multiple-choice evaluation rather than open-ended generation.
HiCUPID~\cite{mok-etal-2025-exploring} evaluates profile-conditioned generation where user preferences and profile attributes are explicitly embedded in the conversation history, reducing the need for implicit memory reasoning over long-term interactions.

Table~\ref{tab:benchmark_comparison} provides a comparison of {\large\textsc{pragma}} with existing personalized-memory and long-context benchmarks.
Overall, existing benchmarks provide limited evaluation of open-ended personalized guidance grounded in evolving user trajectories.

\section{PRAGMA}

\begin{figure*}[t]
    \centering
    \vspace{-0.4cm}
    \includegraphics[width=\linewidth]{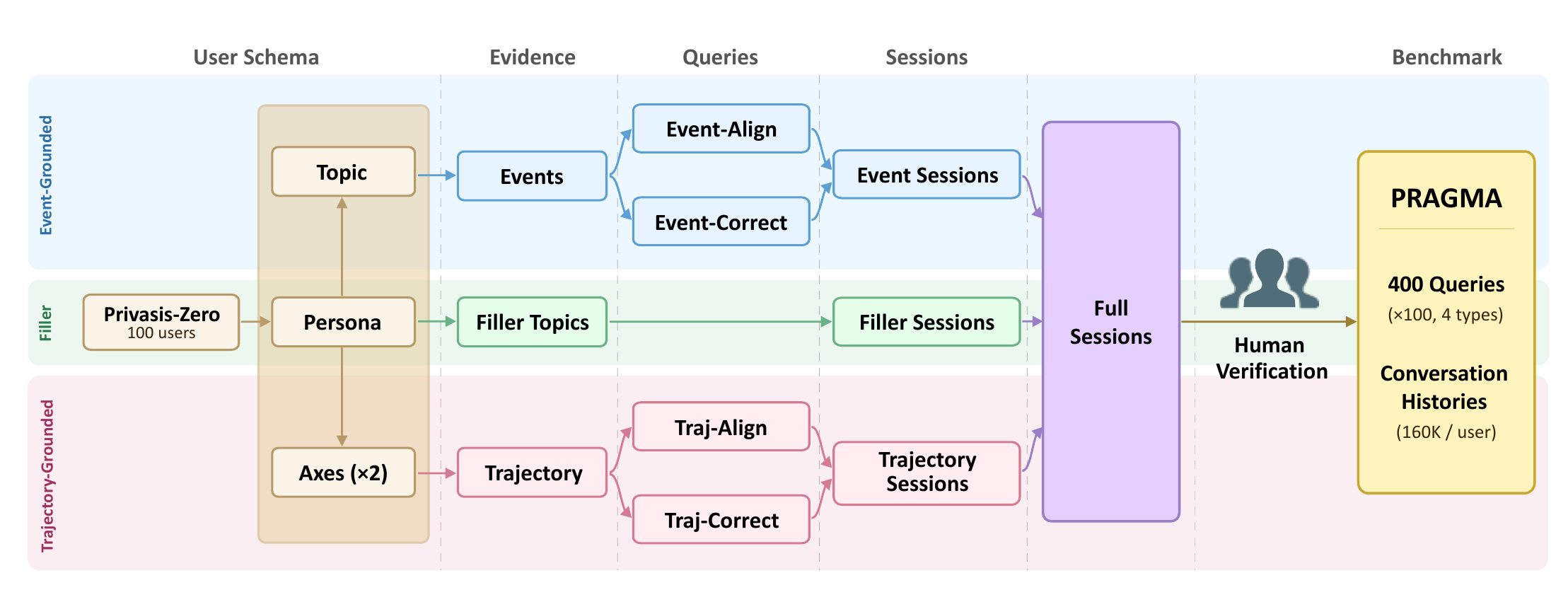}
    \caption{Overview of the {\large{\textsc{pragma}}} benchmark construction pipeline.}
    \vspace{-0.55cm}
    \label{fig:pipeline}
\end{figure*}

We introduce {\large\textsc{pragma}}, a benchmark for evaluating personalized guidance in long-term conversations.
In this section, we describe its construction pipeline, including query design, history generation, evidence annotation, and evaluation protocols.

\subsection{Query Design}


To evaluate real-world personalized guidance, we construct long-term conversational contexts in which user information naturally accumulates over time.
In these settings, users seek open-ended practical guidance, including recommendations, planning support, and decision-making assistance, within ongoing conversations.
However, practical guidance in long-context conversations introduces several challenges.
Relevant context may come from isolated events or emerge across extended interactions, and user requests may be incomplete, outdated, or inconsistent with prior context.
These features are not fully capturable through current factual recall benchmarks with constrained response generation settings.

To thoroughly cover these challenges, we organize queries along two orthogonal dimensions: memory dynamics and query alignment.
Memory dynamics distinguishes between event-level experiences (static) and longitudinal user trajectories (evolving).
Query alignment characterizes whether user assumptions are consistent with prior conversational evidence.
Combining these dimensions yields four query categories (Table~\ref{tab:query_taxonomy}):
\textbf{Event-Align} queries require guidance grounded in coherent event-level experiences;
\textbf{Event-Correct} queries require correcting incorrect event-level assumptions before providing appropriate guidance;
\textbf{Traj-Align} queries require reasoning over evolving user trajectories;
\textbf{Traj-Correct} queries require recognizing when a user's proposed decision conflicts with their longitudinal trajectory and providing corrective guidance.
Example queries for each query type are provided in Appendix~\ref{appx:query_example}.

\begin{table}[t]
\centering
\small
\setlength{\tabcolsep}{3pt}
\renewcommand{\arraystretch}{1.2}

\begin{tabularx}{\columnwidth}{
l
>{\centering\arraybackslash}X
>{\centering\arraybackslash}X
}

\toprule
 & \textbf{Memory-Aligned} & \textbf{Memory-Misaligned} \\
\midrule
\textbf{Static}   & Event-Align & Event-Correct \\
\textbf{Evolving} & Traj-Align  & Traj-Correct  \\
\bottomrule
\end{tabularx}

\caption{{\large{\textsc{pragma}}} query taxonomy.}
\label{tab:query_taxonomy}
\end{table}

\begin{table*}[t]
\centering
\small
\setlength{\tabcolsep}{4.5pt}
\renewcommand{\arraystretch}{1.0}

\begin{tabular}{l l c c c c c c c c}
\toprule

\multicolumn{2}{c}{\textbf{Method}}
& \multicolumn{2}{c}{\makecell{\textbf{Event-Align}\\$(n=100)$}}
& \multicolumn{2}{c}{\makecell{\textbf{Event-Correct}\\$(n=100)$}}
& \multicolumn{2}{c}{\makecell{\textbf{Traj-Align}\\$(n=100)$}}
& \multicolumn{2}{c}{\makecell{\textbf{Traj-Correct}\\$(n=100)$}} \\

\cmidrule(lr){3-4}
\cmidrule(lr){5-6}
\cmidrule(lr){7-8}
\cmidrule(lr){9-10}

& & \textbf{Aln.} & \textbf{Grd.}
& \textbf{Aln.} & \textbf{Grd.}
& \textbf{Aln.} & \textbf{Grd.}
& \textbf{Aln.} & \textbf{Grd.} \\

\midrule

\rowcolor[HTML]{CCDFF0}
\multicolumn{10}{c}{\textbf{GPT-5-mini}} \\
\midrule

\rowcolor[HTML]{EAF4FB}
\textit{Full Context} & & 67.00 & 36.00 & 1.00 & 10.62 & 70.00 & 8.00 & 19.00 & 13.27 \\

\midrule

\rowcolor[HTML]{EAF4FB}
\textit{Dense} & S & 90.00 & 57.00 & 3.00 & 15.56 & 73.00 & 13.00 & 32.00 & 17.68 \\
\rowcolor[HTML]{EAF4FB}
& T & \textbf{98.00} & 52.00 & 2.00 & \textbf{25.18} & 53.00 & 6.00 & \textbf{42.00} & 16.46 \\

\rowcolor[HTML]{EAF4FB}
\textit{Window} & S & 96.00 & 59.00 & 3.00 & 18.50 & 75.00 & 11.00 & 32.00 & 17.06 \\
\rowcolor[HTML]{EAF4FB}
& T & 93.00 & 44.00 & 5.00 & 16.25 & 60.00 & 6.00 & 30.00 & 13.77 \\

\rowcolor[HTML]{EAF4FB}
\textit{BM25} & S & 96.00 & \textbf{60.00} & 4.00 & 13.90 & 64.00 & 10.00 & 28.00 & 16.97 \\
\rowcolor[HTML]{EAF4FB}
& T & 89.00 & 42.00 & 6.00 & 16.72 & 57.00 & 6.00 & 25.00 & 17.46 \\

\midrule

\rowcolor[HTML]{EAF4FB}
\textit{A-MEM} & & 97.00 & 59.00 & \textbf{8.00} & 22.52 & \textbf{81.00} & \textbf{16.00} & 36.00 & \textbf{20.03} \\
\rowcolor[HTML]{EAF4FB}
\textit{Mem0} & & 96.00 & 58.00 & 3.00 & 24.32 & 60.00 & 5.00 & 34.00 & 19.41 \\
\rowcolor[HTML]{EAF4FB}
\textit{SimpleMem} & & 94.00 & \textbf{60.00} & 4.00 & 20.30 & 50.00 & 3.00 & 22.00 & 16.36 \\

\midrule
\midrule

\rowcolor[HTML]{C8E6C7}
\multicolumn{10}{c}{\textbf{Qwen3-30B-A3B-Instruct-2507}} \\
\midrule

\rowcolor[HTML]{E7F5E6}
\textit{Full Context} & & 44.00 & 12.00 & 0.00 & 12.32 & 31.00 & 0.00 & 8.00 & 8.21 \\

\midrule

\rowcolor[HTML]{E7F5E6}
\textit{Dense} & S & 81.00 & 55.00 & 4.00 & 19.13 & 63.00 & 19.00 & \textbf{43.00} & 22.30 \\
\rowcolor[HTML]{E7F5E6}
& T & 82.00 & 44.00 & 8.00 & 22.42 & 45.00 & 7.00 & 38.00 & \textbf{23.52} \\

\rowcolor[HTML]{E7F5E6}
\textit{Window} & S & 84.00 & \textbf{70.00} & 5.00 & 24.33 & 65.00 & \textbf{20.00} & 42.00 & 22.12 \\
\rowcolor[HTML]{E7F5E6}
& T & \textbf{85.00} & 59.00 & 10.00 & \textbf{30.50} & 52.00 & 13.00 & 41.00 & 18.08 \\

\rowcolor[HTML]{E7F5E6}
\textit{BM25} & S & 76.00 & 51.00 & 6.00 & 24.58 & \textbf{68.00} & 14.00 & 30.00 & 16.99 \\
\rowcolor[HTML]{E7F5E6}
& T & 64.00 & 41.00 & 3.00 & 20.05 & 51.00 & 7.00 & 24.00 & 16.42 \\

\midrule

\rowcolor[HTML]{E7F5E6}
\textit{A-MEM} & & 76.00 & 59.00 & 7.00 & 22.19 & 56.00 & 10.00 & 36.00 & 21.33 \\
\rowcolor[HTML]{E7F5E6}
\textit{Mem0} & & 80.00 & 54.00 & \textbf{13.00} & 28.51 & 42.00 & 9.00 & 41.00 & 23.33 \\
\rowcolor[HTML]{E7F5E6}
\textit{SimpleMem} & & 73.00 & 45.00 & 7.00 & 26.38 & 51.00 & 10.00 & 27.00 & 23.28 \\

\bottomrule
\end{tabular}

\caption{
Performance across query types.
Aln.\ denotes \textit{alignment} and Grd.\ denotes \textit{grounding}. S and T indicate \textit{session-} and \textit{turn-level} retrieval.
Bold indicates the best result among practical systems.
}

\vspace{-0.35cm}
\label{tab:main}
\end{table*}

\subsection{Benchmark Construction}
{\large\textsc{pragma}} is built through a controlled generation pipeline with validation to create personalized guidance queries while minimizing shortcut strategies such as recency heuristics and lexical matching.
Additional construction details and examples are provided in Appendix~\ref{appx:bench_details}.

\textbf{User schema design.}
To ensure diversity in user personas and longitudinal behaviors, we begin with 100 synthetic users from Privasis-Zero~\cite{Kim2026PrivasisST}, which provides rich profile information suitable for generating coherent persona-grounded attributes and experiences.
We extract only the profile information relevant for controllable generation, including demographic attributes (e.g., age, income class, native language, citizenship) and user event lists.
Using \texttt{gpt-5}, we generate for each user:
(1) a one-sentence persona summary,
(2) a topic for event-grounded experiences,
and (3) two latent behavioral axes for trajectory-grounded evolution.

The topic and axes are prompted to remain persona-compatible while mutually independent. 

\textbf{Event-grounded query construction.}
For event-grounded queries, we first generate 2--5 timestamped user events associated with the sampled topic over a one-year period (\texttt{2025-05-01} to 
\texttt{2026-04-30}). 
These events are then used to construct two query types corresponding to aligned and corrective guidance settings.

Event-aligned queries are designed to refer to the topic, providing a retrieval cue while still requiring the model to interpret how previous experiences should influence the recommendation.
Event-corrective queries intentionally contain mixed or partially incorrect recollections constructed from multiple prior events. 
Rather than directly requesting factual correction, the user asks for practical guidance based on a mistaken premise, requiring models to identify inconsistencies in the user's assumptions before producing appropriate guidance.

\textbf{Trajectory-grounded query construction.}
For trajectory-grounded queries, we generate longitudinal user trajectories consisting of 4--8 timestamped states over two independent behavioral axes.
We intentionally construct trajectories over multiple simultaneously evolving behavioral axes.
When only a single attribute changes, the task can often collapse into retrieving the user's most recent state, whereas multi-axis trajectories require models to jointly track multiple aspects of the user over time.

Using these trajectories, we construct two types of trajectory-grounded guidance queries. 
For trajectory-aligned queries, users explicitly refer to the underlying axes as qualities they currently value or wish to emphasize. 
To avoid direct lexical shortcuts, we rewrite the axes into abstract descriptors (e.g., single-word concepts) before they appear in the query, requiring models to infer the underlying trajectory from the conversational history.

For trajectory-corrective queries, we generate decisions that appear individually plausible but conflict with the user's longer-term trajectory.

\textbf{Conversation history construction.}
Each evidence item is expanded into a natural conversational session using \texttt{gpt-5-mini}. 
To simulate realistic long-term interactions and increase retrieval difficulty, we additionally generate 30 filler topics per user that are unrelated to the target topic and trajectory axes while remaining consistent with the user's persona. 
These filler topics are similarly expanded into conversational sessions.

Evidence and filler sessions are then concatenated into a single long-term conversation history. 
Evidence sessions are first ordered according to their timestamps, after which filler sessions are uniformly inserted between them to avoid positional concentration. 
In particular, the first and last sessions are always filler sessions, preventing trivial boundary-position or recency heuristics.

The final benchmark contains 100 users and 400 queries (four query types per user). 
Each user history contains approximately 160K conversational tokens shared across the four associated queries.

\textbf{Human validation.}
All generated queries and evidence annotations are manually reviewed before finalizing. 
Human validation is particularly important for trajectory-grounded and corrective queries, where subtle inconsistencies or unintended shortcuts can significantly reduce benchmark difficulty. Annotators were instructed to verify query realism, evidence correctness, and consistency with the intended query types and reasoning requirements.
Given the complexity of this validation over long conversational histories and distributed evidence, we prioritize rigorous instance-level quality over increasing benchmark scale.
Additional validation details and benchmark statistics are provided in Appendix~\ref{appx:human} and~\ref{appx:bench_stats}.

\subsection{Annotations and Evaluation Protocols}

{\large{\textsc{pragma}}} comprehensively evaluates both retrieval- and response-level performance.

\textbf{Retrieval Evaluation.}
For the retrieval evaluation, each query is annotated with the necessary evidence sessions to generate a response. 
These annotations enable standard retrieval-based evaluation using metrics such as Recall, Precision, F1, and Exact Recall at the session level.
Because many queries require integrating information distributed across multiple sessions, session-level evidence coverage serves as the primary retrieval metric.

\textbf{Response Evaluation and Metadata.}
We evaluate generated responses using \texttt{gpt-5} as an LLM judge with query-type-specific rubrics along two dimensions:
(1) \textbf{Alignment}, measuring consistency with the user’s experiences and longitudinal history, and 
(2) \textbf{Grounding}, measuring explicit support from the annotated evidence.

Each of the four query types has separate alignment and grounding rubrics, yielding eight rubric sets. Detailed rubrics and judging prompts are provided in Appendix~\ref{appx:eval_prompt}.
The benchmark also provides query types, annotated evidence sessions, gold responses, and no-context responses.
Gold responses use summarized evidence and evaluation rubrics, while no-context responses use only the query, both generated with \texttt{gpt-5}.

\section{Experimental Setup}

\begin{table*}[t]
\centering
\small
\setlength{\tabcolsep}{6.5pt}
\renewcommand{\arraystretch}{0.92}

\begin{tabular}{l l c c c c c c c c}
\toprule

\multicolumn{2}{c}{\textbf{Method}}
& \multicolumn{2}{c}{\makecell{\textbf{Event-Align}\\$(n=100)$}}
& \multicolumn{2}{c}{\makecell{\textbf{Event-Correct}\\$(n=100)$}}
& \multicolumn{2}{c}{\makecell{\textbf{Traj-Align}\\$(n=100)$}}
& \multicolumn{2}{c}{\makecell{\textbf{Traj-Correct}\\$(n=100)$}} \\

\cmidrule(lr){3-4}
\cmidrule(lr){5-6}
\cmidrule(lr){7-8}
\cmidrule(lr){9-10}

& & \textbf{Aln.} & \textbf{Grd.}
& \textbf{Aln.} & \textbf{Grd.}
& \textbf{Aln.} & \textbf{Grd.}
& \textbf{Aln.} & \textbf{Grd.} \\

\midrule

\rowcolor[HTML]{CCDFF0}
\multicolumn{10}{c}{\textbf{GPT-5-mini}} \\
\midrule

\rowcolor[HTML]{EAF4FB}
\textit{No-Context}
& & 41.00 & 0.00
& 2.00 & 6.85
& 4.00 & 0.00
& 0.00 & 2.49 \\

\rowcolor[HTML]{EAF4FB}
\textit{Oracle Session}
& & 84.00 & 39.00
& 4.00 & 17.12
& 78.00 & 17.00
& 40.00 & 25.18 \\

\rowcolor[HTML]{EAF4FB}
\textit{Oracle Summary}
& & \textbf{99.00} & \textbf{97.00}
& \textbf{12.00} & \textbf{45.62}
& \textbf{99.00} & \textbf{86.00}
& \textbf{67.00} & \textbf{71.76} \\

\midrule
\midrule

\rowcolor[HTML]{C8E6C7}
\multicolumn{10}{c}{\textbf{Qwen3-30B-A3B-Instruct-2507}} \\
\midrule

\rowcolor[HTML]{E7F5E6}
\textit{No-Context}
& & 39.00 & 0.00
& 1.00 & 7.94
& 0.00 & 0.00
& 1.00 & 1.68 \\

\rowcolor[HTML]{E7F5E6}
\textit{Oracle Session}
& & 74.00 & 48.00
& 2.00 & 15.14
& 73.00 & 26.00
& 35.00 & 18.47 \\

\rowcolor[HTML]{E7F5E6}
\textit{Oracle Summary}
& & \textbf{84.00} & \textbf{66.00}
& \textbf{17.00} & \textbf{32.16}
& \textbf{100.00} & \textbf{97.00}
& \textbf{71.00} & \textbf{60.39} \\

\bottomrule
\end{tabular}

\caption{
No-Context and oracle settings.
Aln.\ denotes \textit{alignment} and Grd.\ denotes \textit{grounding}.
}
\vspace{-0.35cm}

\label{tab:main_reference}
\end{table*}

\subsection{Models and Baselines}

We evaluate RAG and memory systems using two generation models: \texttt{gpt-5-mini}~\cite{singh2026openaigpt5card} and \texttt{qwen3-30b-a3b-instruct}~\cite{yang2025qwen3technicalreport}. 
All methods use \texttt{bge-base-en-v1.5}~\cite{xiao2024cpackpackedresourcesgeneral} embeddings.
We include three reference conditions: 
(1) \textbf{No-context}, where the model answers using only the query without conversational history; 
(2) \textbf{Oracle-session}, where the gold evidence sessions are directly provided; and 
(3) \textbf{Oracle-summary}, where the model receives summarized evidence from annotated metadata. 
As a long-context baseline, we evaluate \textbf{full-context}, where the model receives the complete conversational history.
For retrieval-based baselines, we evaluate RAG systems under both turn-level and session-level retrieval settings.
We compare \textbf{dense} retrieval, \textbf{BM25} sparse retrieval, and a simple \textbf{window} retrieval strategy that augments retrieved turns with nearby conversational context.

We further evaluate representative memory systems for long-term conversational personalization, including \textbf{A-MEM}~\cite{Xu2025AMEMAM}, \textbf{Mem0}~\cite{Chhikara2025Mem0BP}, and \textbf{SimpleMem}~\cite{Liu2026SimpleMemEL}, which differ in how conversational histories are stored and retrieved.
Each system provides the top-$k$ retrieved memory records as context for response generation.
Unless otherwise specified, all methods use comparable retrieval budgets and the same backbone model for memory ingestion.
Additional implementation details are provided in Appendix~\ref{appx:config}.

\subsection{Evaluation Metrics}

\textbf{Retrieval Evaluation.}
Using the annotated evidence sessions, we evaluate whether systems retrieve required information for each query.
Because annotations are provided only at the session level, retrieval evaluation is reported only for session-level RAG systems.
We report \textbf{Recall}, measuring the fraction of annotated evidence sessions retrieved, and \textbf{Exact Recall}, measuring whether all required evidence sessions are retrieved.

\textbf{Response Evaluation.}
We evaluate generated responses using \texttt{gpt-5} with the query-type-specific alignment and grounding rubrics provided in {\large{\textsc{pragma}}}.
To support evaluation robustness, we report evaluations using \texttt{gemini-3.1-pro-preview}~\cite{gemini31propreview} and \texttt{claude-opus-4.6}~\cite{claudeopus46} with inter-judge agreement results in Appendix~\ref{appx:gemini}.

\begin{table*}[t]
\centering
\footnotesize
\setlength{\tabcolsep}{2.8pt}
\renewcommand{\arraystretch}{1.1}

\begin{tabular}{llcccc|cccc|cccc|cccc}
\toprule

\multicolumn{2}{l}{\textbf{Method}}
& \multicolumn{4}{c|}{\makecell{\textbf{Event-Align}\\$(n=100)$}}
& \multicolumn{4}{c|}{\makecell{\textbf{Event-Correct}\\$(n=100)$}}
& \multicolumn{4}{c|}{\makecell{\textbf{Traj-Align}\\$(n=100)$}}
& \multicolumn{4}{c}{\makecell{\textbf{Traj-Correct}\\$(n=100)$}} \\

& & \textbf{Rec.} & \textbf{Ex.} & \textbf{Aln.} & \textbf{Grd.}
& \textbf{Rec.} & \textbf{Ex.} & \textbf{Aln.} & \textbf{Grd.}
& \textbf{Rec.} & \textbf{Ex.} & \textbf{Aln.} & \textbf{Grd.}
& \textbf{Rec.} & \textbf{Ex.} & \textbf{Aln.} & \textbf{Grd.} \\

\midrule

\textit{Dense} & S
    & 85.30 & 50.00 & 81.00 & 55.00
    & 89.07 & \textbf{68.00} & 4.00 & 19.13
    & 62.48 & 0.00 & 63.00 & 19.00
    & \textbf{69.83} & \textbf{18.00} & \textbf{43.00} & 22.30 \\
    & T
    & -- & -- & 82.00 & 44.00
    & -- & -- & 8.00 & 22.42
    & -- & -- & 45.00 & 7.00
    & -- & -- & 38.00 & \textbf{23.52} \\

\textit{BM25} & S
    & 57.45 & 7.00 & 76.00 & 51.00
    & 88.70 & 63.00 & 6.00 & 24.58
    & 51.95 & 0.00 & 68.00 & 14.00
    & 56.45 & 6.00 & 30.00 & 16.99 \\
    & T
    & -- & -- & 64.00 & 41.00
    & -- & -- & 3.00 & 20.05
    & -- & -- & 51.00 & 7.00
    & -- & -- & 24.00 & 16.42 \\

\textit{Window} & S
    & 73.45 & 18.00 & 84.00 & \textbf{70.00}
    & 84.40 & 59.00 & 5.00 & 24.33
    & 56.67 & 2.00 & 65.00 & 20.00
    & 53.70 & 8.00 & 42.00 & 22.12 \\
    & T
    & -- & -- & 85.00 & 59.00
    & -- & -- & \textbf{10.00} & \textbf{30.50}
    & -- & -- & 52.00 & 13.00
    & -- & -- & 41.00 & 18.08 \\

\midrule

\textit{Dynamic} & S
    & 69.20 & 16.00 & 74.00 & 48.00
    & 77.57 & 40.00 & 5.00 & 17.99
    & 64.17 & \textbf{4.00} & 40.00 & 8.00
    & 64.15 & 12.00 & 30.00 & 15.12 \\
    & T
    & -- & -- & 67.00 & 25.00
    & -- & -- & 4.00 & 18.83
    & -- & -- & 11.00 & 3.00
    & -- & -- & 19.00 & 13.27 \\


\textit{QR} & S
      & \textbf{86.75} & \textbf{51.00} & 83.00 & 57.00
      & \textbf{89.27} & 67.00 & 7.00 & 21.93
      & \textbf{67.88} & 0.00 & \textbf{74.00} & \textbf{28.00}
      & 69.33 & 16.00 & 41.00 & 18.61 \\
      & T
      & -- & -- & \textbf{86.00} & 49.00
      & -- & -- & 7.00 & 27.71
      & -- & -- & 55.00 & 8.00
      & -- & -- & 32.00 & 19.61 \\

\textit{AdaK} & S
      & 79.50 & 39.00 & 82.00 & 55.00
      & 83.50 & 55.00 & 4.00 & 21.36
      & 59.43 & 0.00 & 63.00 & 17.00
      & 62.46 & 15.00 & 33.00 & 21.26 \\
      & T
      & -- & -- & 74.00 & 38.00
      & -- & -- & 8.00 & 20.93
      & -- & -- & 28.00 & 4.00
      & -- & -- & 28.00 & 16.92 \\

\bottomrule
\end{tabular}

\caption{RAG accuracy across query types and metrics. S and T indicate session- and turn-level retrieval, while QR and AdaK indicate query rewriting and Adaptive-$K$. Rec. and Ex. denote evidence recall and exact recall.}
\vspace{-5pt}

\label{tab:rag}

\end{table*}

\section{Experimental Results}

\subsection{Main Results}

In Table~\ref{tab:main}, Full-Context remains ineffective despite being given complete conversational history across both generation models.
Specifically, under gpt-5-mini, Full-Context achieves only 8.00 grounding on Trajectory-Align, substantially below practical retrieval and memory systems.
These results suggest that long conversational histories alone are insufficient for robust personalized guidance.

Performance also varies substantially across query types.
Trajectory-grounded queries highlight the importance of broader conversational context.
For example, under gpt-5-mini, Dense-session improves Trajectory-Align alignment from 53.00 to 73.00 compared to turn-level retrieval, while A-MEM achieves the strongest performance at alignment.
Nevertheless, grounding performance remains limited across practical systems, highlighting the difficulty of producing well-grounded personalized guidance over evolving user trajectories. Corrective queries are particularly challenging across all systems.
Even when user assumptions conflict with prior memory, models often fail to produce corrective responses.
Under gpt-5-mini, Event-Correct alignment remains between 2.00 and 8.00 across all practical systems, while Trajectory-Correct alignment remains below 43.00.


Across nearly all systems, alignment scores are substantially higher than grounding scores.
For example, with gpt-5-mini, A-MEM achieves 81.00 alignment but only 16.00 grounding on Trajectory-Align, suggesting that models often generate plausible personalized guidance without effectively grounding it in conversational evidence.

We further evaluate additional retrieval and memory baselines, which broadly support our main findings (Appendix~\ref{more_baselines}).

\subsection{Gold Retrieval is Not Enough}

Table~\ref{tab:main_reference} presents no-context lower bounds and oracle settings for personalized guidance. Across both generation models, No-Context achieves near-zero grounding and very low alignment, indicating that PRAGMA instances are not solvable from query-only priors or generic guidance patterns. In other words, personalized guidance fundamentally must utilize the conversational memory.

Furthermore, to separate retrieval failure from evidence utilization failure, we evaluate two increasingly model-friendly oracle settings: Oracle-Session directly provides the annotated gold evidence sessions, while Oracle-Summary further compresses the same evidence into concise summaries containing the key information needed for guidance. Despite removing retrieval as a bottleneck in both settings, we observe a substantial gap between Oracle-Session and Oracle-Summary, particularly on trajectory-grounded and corrective queries. For example, under gpt-5-mini, Oracle-Summary achieves 99.00 alignment and 86.00 grounding on Trajectory-Align, whereas Oracle-Session reaches only 78.00 and 17.00, respectively. This gap suggests that merely exposing the relevant conversation sessions is insufficient; models still struggle to organize and synthesize longitudinal evidence unless it is presented in an explicitly distilled form. 
Notably, even Oracle-Summary remains far from perfect on corrective queries, suggesting that these challenges cannot be resolved through retrieval quality alone.

\begin{table*}[h]
\centering
\small
\setlength{\tabcolsep}{4pt}
\begin{tabular}{llcccccc}
\toprule
\multicolumn{2}{l}{\textbf{Method}} & \textbf{Recall} & \textbf{Exact} & \textbf{Alignment} & \makecell{\textbf{Alignment}\\\textbf{(2-Stage)}} & \textbf{Grounding} & \makecell{\textbf{Grounding}\\\textbf{(2-Stage)}} \\
\midrule
\textit{Dense} & S & \textbf{89.07} & \textbf{68.00} & 76.0 & 99.0 (+23.0) & 49.3 & 74.3 (+25.0) \\
               & T    & -- & -- & 67.0 & 95.0 (+28.0) & 49.3 & 71.1 (+21.8) \\
\textit{Window} & S & 84.40 & 59.00 & 77.0 & \textbf{100.0 (+23.0)} & 55.9 & \textbf{80.2 (+24.3)} \\
                & T    & -- & -- & 77.0 & 99.0 (+22.0) & 54.2 & 78.0 (+23.9) \\
\textit{BM25} & S & 88.70 & 63.00 & 68.0 & \textbf{100.0 (+32.0)} & 48.0 & 79.5 (+31.5) \\
              & T    & -- & -- & 67.0 & 97.0 (+30.0) & 41.2 & 63.1 (+21.9) \\
\midrule
\textit{A-MEM}     & & -- & -- & \textbf{80.0} & \textbf{100.0 (+20.0)} & \textbf{57.2} & 77.9 (+20.7) \\
\textit{Mem0}      & & -- & -- & 60.0 & 87.0 (+27.0) & 39.8 & 59.0 (+19.2) \\
\textit{SimpleMem} & & -- & -- & 70.0 & 93.0 (+23.0) & 45.8 & 62.3 (+16.5) \\
\bottomrule
\end{tabular}
\caption{Results on corrective guidance queries involving misaligned user assumptions. S and T indicate session- and turn-level retrieval. Alignment and Grounding denote one-stage generation with an explicit instruction that the user’s assumption may be incorrect; 2-Stage first identifies inconsistencies before generating the response.}
\vspace{-0.35cm}
\label{tab:antisyco}
\end{table*}

We observe the same pattern across generation models of varying scales: Oracle-Summary consistently outperforms Oracle-Session, further showing that gold evidence access alone does not ensure effective memory utilization (Appendix~\ref{more_models}).

\subsection{Retrieval-Response Discrepancy}

To further analyze the gap between retrieval and response, we evaluate several RAG variants adapted to our setting, including Dynamic Retrieval~\cite{Jiang2023ActiveRA}, Query Rewriting~\cite{Gao2022PreciseZD}, and Adaptive-$k$~\cite{Taguchi2025EfficientCS} retrieval.
All methods use the same embedding model and retrieval budget ($k\leq10$), with \texttt{qwen3-30b-a3b-instruct} for generation.

Table~\ref{tab:rag} shows that many methods fail to recover the complete evidence required for personalized guidance, particularly for trajectory-grounded and corrective queries.
For example, Dense retrieval achieves 62.48 Recall but 0.00 Exact Recall on Trajectory-Align, while BM25 reaches 51.95 Recall with similarly low complete evidence recovery.

At the same time, strong retrieval performance does not necessarily lead to strong downstream responses.
This pattern is particularly clear on Event-Correct: despite high recall (88.70–89.27) and exact recall (63.00–68.00), Dense, BM25, and Query Rewriting all exhibit low alignment (4.00–7.00) and grounding (19.13–24.58).
Overall, these results suggest that personalized guidance requires improvements in both conversational retrieval and downstream memory utilization.

We also observe clear differences between retrieval granularities.
For trajectory-grounded queries, session-level retrieval consistently outperforms turn-level retrieval on alignment.
On Trajectory-Align, Dense improves from 45.00 to 63.00 and BM25 from 51.00 to 68.00.
In contrast, Event-Correct queries often favor turn-level retrieval, as excessive session-level context can obscure fine-grained corrective evidence.
For instance, Window grounding improves from 24.33 to 30.50 with turn-level retrieval.
Overall, the results suggest that improving retrieval alone is insufficient for robust personalized guidance.

\section{Analysis}

\subsection{Models Still Fail at Grounding}

To better understand corrective guidance failures, we analyze Event-Correct queries, where user assumptions conflict with conversational history.
These queries require identifying inconsistencies and generating evidence-grounded corrections.

Adding an instruction that the user’s assumption may be incorrect substantially improves alignment across systems, suggesting that inconsistency detection itself is not the primary bottleneck. 
We further evaluate a two-stage setup that first identifies inconsistencies and then generates a corrective response conditioned on them. 
As shown in Table~\ref{tab:antisyco}, alignment improves dramatically under this setup, often reaching near-perfect performance.

However, grounding remains substantially lower despite strong retrieval performance and explicit inconsistency identification. 
These results suggest that corrective guidance failures cannot be explained solely by inconsistency detection failures; reliably recovering and utilizing the appropriate conversational evidence remains challenging.
Such grounding failures can propagate across future interactions, where earlier responses themselves become part of the conversational context used for subsequent reasoning. 
A detailed follow-up case study is provided in Appendix~\ref{appx:follow-up}.

The extremely low corrective alignment observed across practical systems also suggests a broader tendency toward over-accommodation to user assumptions, consistent with prior observations of sycophantic behavior in instruction-tuned LLMs~\cite{DBLP:conf/iclr/SharmaTKDABDHJK24, Hong2025MeasuringSO}.
In personalized guidance settings, this behavior becomes particularly problematic because effective assistance may require challenging the user’s current belief rather than simply validating it.

\subsection{Where Do Memories Get Lost?}


Previous experiments suggest that retrieval alone is insufficient for personalized guidance.
To better understand system failures, we decompose the memory pipeline into three aspects:
\textbf{memory preservation}, whether evidence remains preserved in memory;
\textbf{retrieval accessibility}, whether preserved evidence is successfully retrieved;
and \textbf{response utilization}, whether retrieved evidence is reflected in the final response.
All evaluations use entailment-style LLM judgments with \texttt{gpt-5-nano}.

Table~\ref{tab:memsys} reveals substantial tradeoffs across memory systems. 
A-MEM achieves nearly perfect preservation across all query types by storing raw conversational content, while Mem0 and SimpleMem lose information during memory rewriting and compression. 
However, strong preservation does not necessarily translate into downstream utilization; on Traj-Correct, A-MEM preserves 99.1\% of evidence but retrieves only 48.9\%.

\begin{table}[t]
\centering
\small
\setlength{\tabcolsep}{5pt}

\begin{tabular}{llccc}
\toprule
\textbf{Type} & \textbf{System} & \textbf{St} & \textbf{Rt} & \textbf{Rs} \\
\midrule

\multirow{3}{*}{EA}
& A-MEM    & \textbf{98.2} & \textbf{60.4} & 70.3 \\
& Mem0     & 82.6 & 23.0 & \textbf{78.7} \\
& SimpleMem& 64.2 & 41.3 & 74.8 \\

\midrule

\multirow{3}{*}{EC}
& A-MEM    & \textbf{98.0} & \textbf{81.2} & 66.4 \\
& Mem0     & 83.2 & 57.7 & 62.7 \\
& SimpleMem& 65.1 & 65.5 & \textbf{72.0} \\

\midrule

\multirow{3}{*}{TA}
& A-MEM    & \textbf{99.4} & \textbf{49.6} & 81.6 \\
& Mem0     & 78.9 & 17.1 & 81.3 \\
& SimpleMem& 62.1 & 20.6 & \textbf{82.6} \\

\midrule

\multirow{3}{*}{TC}
& A-MEM    & \textbf{99.1} & \textbf{48.9} & 63.3 \\
& Mem0     & 77.5 & 20.1 & \textbf{81.2} \\
& SimpleMem& 58.1 & 31.0 & 72.5 \\

\bottomrule
\end{tabular}

\caption{
Evidence preservation across memory stages. St: fraction of gold evidence preserved in storage; Rt: fraction of stored evidence successfully retrieved; Rs: fraction of retrieved evidence reflected in the response.
}
\label{tab:memsys}
\end{table}

\begin{table}[t]
\centering
\small
\setlength{\tabcolsep}{8pt}
\renewcommand{\arraystretch}{1.1}

\begin{tabular}{lcccc}
\toprule
\textbf{System} & \textbf{EA} & \textbf{EC} & \textbf{TA} & \textbf{TC} \\
\midrule
\textit{A-MEM} & 70.3 & 66.4 & 81.6 & 63.3 \\
\textit{Mem0} & \textbf{78.7} & 62.7 & 81.3 & \textbf{81.2} \\
\textit{SimpleMem} & 74.8 & \textbf{72.0} & \textbf{82.6} & 72.5 \\
\midrule
\textit{Dense} & 67.0 & 61.2 & 76.8 & 51.5 \\
\textit{Window} & 66.0 & 61.0 & 74.2 & 55.6 \\
\textit{BM25} & 68.9 & 60.7 & 75.6 & 53.8 \\
\bottomrule
\end{tabular}
\vspace{-1mm}
\caption{Response-stage evidence utilization across query types.
EA: Event-Align, EC: Event-Correct, TA: Trajectory-Align, TC: Trajectory-Correct.}
\vspace{0.2cm}
\label{tab:memsys_response}
\end{table}

We further compare response-stage utilization across memory systems and RAG pipelines in Table~\ref{tab:memsys_response}.
Despite preserving less information overall, summarized memory systems often achieve stronger downstream utilization than more detailed memory representations and standard RAG. This suggests that memory abstraction is not merely a compression mechanism, but a critical interface between retrieval and generation: concise structured memories may discard some low-level conversational detail, yet expose the remaining evidence in a form that generation models can more reliably incorporate into personalized guidance. In contrast, raw conversational context can preserve more evidence while still leaving the generator to identify, organize, and synthesize the relevant implications. Overall, current systems struggle to simultaneously optimize preservation, retrieval accessibility, and utilization, highlighting the need for memory architectures that store information not only accurately, but also in generation-usable forms.


\section{Conclusion}

We introduced {\large\textsc{pragma}}, a benchmark for evaluating personalized guidance in long-term conversations beyond factual recall.
It evaluates whether models can provide grounded guidance under evolving user preferences and potentially incorrect assumptions.
Experiments across RAG, memory systems, and long-context models reveal substantial failures in both memory retrieval and utilization.
Even when relevant evidence is retrieved, models often fail to generate grounded personalized guidance, highlighting the need for memory systems that support robust longitudinal reasoning and memory-grounded generation beyond retrieval.

\section*{Limitations}

{\large\textsc{pragma}} focuses on controlled evaluation of memory-grounded personalized guidance, and several limitations remain for future work.
First, although the benchmark is human-validated, the conversational histories are generated through a controllable synthetic pipeline.
This design enables evidence annotation, trajectory control, and systematic evaluation across diverse memory scenarios, but may not fully capture the ambiguity and variability of natural long-term human conversations.

Second, the benchmark evaluates guidance generation in a single-turn setting.
In real deployments, conversational agents may recover from incomplete memory retrieval through iterative interaction or follow-up dialogues.
Future works could extend {\large\textsc{pragma}} toward interactive multi-turn evaluation of memory utilization and conversational recovery.

Finally, retrieval evaluation is based on annotated evidence sessions rather than fine-grained reasoning traces.
While this abstraction improves annotation reliability and scalability, some queries may admit multiple valid reasoning paths or rely on partially implicit evidence distributed across conversations.
Developing more fine-grained evaluation protocols for longitudinal memory reasoning remains an important direction for future research.

\section*{Ethical Considerations}
{\large\textsc{pragma}} is constructed from fully synthetic conversational histories and does not contain real user data or personally identifiable information.
However, models may overfit to benchmark-specific annotation structures or reasoning patterns rather than developing robust long-term personalization capabilities.
In addition, although {\large\textsc{pragma}} is designed to cover diverse personas and longitudinal behaviors, synthetic generation pipelines may still underrepresent certain cultural, linguistic, or interactional patterns, potentially introducing unintended biases in evaluation outcomes.
We therefore encourage future work to evaluate whether improvements on {\large\textsc{pragma}} transfer to more open-ended and realistic conversational settings.

\section*{Acknowledgments}
This work was supported by the IITP(Institute of Information \& Communications Technology Planning \& Evaluation)-ITRC(Information Technology Research Center) grant funded by the Korea government(Ministry of Science and ICT)(IITP-2025-RS-2024-00437633).
This work was conducted in collaboration with LYWAY on domain-specific AI research, whose support for our memory research and funding of the API costs we gratefully acknowledge.
K. Jung is with ASRI, Seoul National University, Korea.
The Institute of Engineering Research at Seoul National University provided research facilities for this work.


\clearpage
\bibliography{custom}

\clearpage

\appendix
\section{Benchmark Construction Details}
\label{appx:bench_details}

\subsection{Prompt Templates and Generation Pipeline}
Below, we provide abbreviated prompt templates used in our benchmark generation pipeline.

\vspace{7pt}

\noindent

\begin{minipage}{\linewidth}
\promptgroup{Event Generation}{%
Given the persona, axes, and topic, create a few specific past events for the user.

\textbf{Events}: focus on the topic; should NOT be relevant to the axes; include temporal or spatial words; start with ``The user.''\newline
\textbf{Timestamps}: between 2025-05-01 and 2026-04-30, sorted in temporal order.
}
\vspace{2pt}
\captionof{figure}{Prompt template for event generation.}
\label{fig:prompt_event}
\end{minipage}

\noindent

\begin{minipage}{\linewidth}
\promptgroup{Trajectory Generation}{%
Given the persona and axes, create a trajectory for the user.

\textbf{Trajectory}:
\begin{itemize}[noitemsep, topsep=2pt, leftmargin=12pt]
  \item Sequential with 4--8 states.
  \item Each state relevant to at least one axis; some states relevant to only one axis.
  \item Include the user's initial preference and drift over time.
  \item Vary in pattern (gradual shift, oscillation, plateau, partial reversal).
  \item Avoid overly clean or perfectly structured progression.
  \item The last state should NOT represent the user's current state.
\end{itemize}
\textbf{Timestamps}: between 2025-05-01 and 2026-04-30, sorted in temporal order.
}
\vspace{2pt}
\captionof{figure}{Prompt template for trajectory generation.}
\label{fig:prompt_traj}
\end{minipage}

\noindent

\begin{minipage}{\linewidth}
\promptgroup{Event-Aligned Query Generation (Type 1)}{%
Given the topic and past user events, summarize the events then create a user query.

\textbf{Summary}: first-person (``I''), concise.\newline
\textbf{Query}:
\begin{itemize}[noitemsep, topsep=2pt, leftmargin=12pt]
  \item Should NOT restate or hint at the events; remain implicit.
  \item Should ask for recommendation or advice on the topic.
  \item Should be implicitly grounded in past experiences without meta-phrases such as ``based on my experiences.''
  \item Should sound natural, concise, and underspecified.
\end{itemize}
\textbf{Output format}:\newline
Summary: \textless summary\textgreater\newline
Query: \textless query\textgreater
}
\vspace{2pt}
\captionof{figure}{Prompt template for Event-Aligned query generation. Additional few-shot examples were provided during generation.}
\label{fig:prompt_type1}
\end{minipage}

\noindent
\begin{minipage}{\linewidth}
\promptgroup{Event-Corrective Query Generation}{%
Given the past user events, select multiple events then generate a user query.

\textbf{Query}:
\begin{itemize}[noitemsep, topsep=2pt, leftmargin=12pt]
  \item Construct a plausible but imperfect recollection by blending multiple past events.
  \item Be \textit{confidently incorrect} about the selected events.
  \item Remain grounded in the original events (no completely new activities).
  \item Ask for recommendation or advice based on the incorrect recollection.
  \item Be concise.
\end{itemize}
\textbf{Selected events}: output the indices of the mixed-up events used in the query.
}
\vspace{2pt}
\captionof{figure}{Prompt template for Event-Corrective query generation. Additional few-shot examples were provided during generation.}
\label{fig:prompt_type2}
\end{minipage}

\noindent
\begin{minipage}{\linewidth}
\promptgroup{Trajectory-Aligned Query Generation}{%
Given the persona, rewritten axes, and user trajectory, create a user query.

\textbf{Query}:
\begin{itemize}[noitemsep, topsep=2pt, leftmargin=12pt]
  \item A realistic self-positioning query (e.g., CV, bio, introduction, application).
  \item Ask how to position themselves along the rewritten axes using past experience.
  \item Include only the rewritten axes terms; do NOT state past experiences or trajectory.
  \item Remain underspecified without explicitly stating progression or trade-offs.
  \item Be concise and natural.
\end{itemize}
\textbf{Output format}: Query: \textless query\textgreater
}
\vspace{2pt}
\captionof{figure}{Prompt template for Trajectory-Aligned query generation.}
\label{fig:prompt_type3}
\end{minipage}

\noindent
\begin{minipage}{\linewidth}
\promptgroup{Trajectory-Corrective Query Generation}{%
Given the user's trajectory, select multiple states, then create a user query.

\textbf{Query}:
\begin{itemize}[noitemsep, topsep=2pt, leftmargin=12pt]
  \item Present a concrete situation and a decision the user is considering.
  \item The decision should be reasonable without additional context, but subtly misaligned with the trajectory.
  \item The misalignment should emerge only when reasoning across \textit{both} axes.
  \item Do NOT reference the trajectory, mention trade-offs, or signal doubt.
  \item Be concise and natural.
\end{itemize}
\textbf{Selected states}: output the indices of states that serve as counter-evidence.
}
\vspace{2pt}
\captionof{figure}{Prompt template for Trajectory-Corrective query generation. Additional few-shot examples were provided during generation.}
\label{fig:prompt_type4}
\end{minipage}

\subsection{End-to-End Construction Example}
\label{appx:pipeline_example}
Table~\ref{tab:construction-pipeline-example} provides a running example of the benchmark construction pipeline in Figure~\ref{fig:pipeline}, tracing a single user from the initial Privasis-Zero profile to the final \large{\textsc{pragma}} instances.
The user profile is transformed into a persona, event topic, and trajectory axes, which guide the generation of events, longitudinal trajectories, filler topics, and the four query types.
These components are expanded into timestamped conversational sessions and combined into a shared long-term history, with each query linked to its corresponding evidence and evaluation metadata.

\subsection{Human Validation Guidelines}
\label{appx:human}
We conducted a human validation study to verify that generated conversations, queries, and evidence annotations support the intended personalized-memory reasoning tasks. Validation was conducted by a team of six annotators, including three co-authors, two NLP researchers, and one researcher in linguistics. Annotators were provided with the user metadata, target query, annotated evidence sessions, and the corresponding conversation history.
For each example, annotators answered query-specific validation questions using a binary yes/no rubric, with optional free-form comments and query rewrites for unnatural or ambiguous cases.

Annotators were instructed to reject examples when:
(1) the query was unnatural or unrealistic,
(2) evidence annotations were incomplete or incorrect,
(3) filler sessions leaked relevant information,
(4) the intended inconsistency was weak or unsupported, or
(5) the query could be solved through superficial heuristics without reasoning over the provided history.

Type-specific validation criteria included:

\begin{itemize}
    \item \textbf{Event-Aligned:}
    Whether the query required event-grounded personalized guidance and whether irrelevant filler sessions remained unrelated to the target topic.

    \item \textbf{Event-Corrective:}
    Whether the query introduced a realistic event-level misconception requiring corrective guidance and whether supporting evidence was naturally distributed across sessions.

    \item \textbf{Trajectory-Aligned:}
    Whether answering the query required reasoning over longitudinal changes in the user's preferences, goals, or circumstances.

    \item \textbf{Trajectory-Corrective:}
    Whether the proposed user decision meaningfully conflicted with the established trajectory and required corrective reasoning grounded in the conversational history.
\end{itemize}

When a query was understandable but unnatural, annotators were encouraged to provide rewritten versions while preserving the intended query type and evidence dependency.
All annotators were compensated based on estimated task completion time in accordance with local institutional research assistant compensation practices.
Because the validation process was designed primarily for quality control and iterative refinement, examples were divided across annotators rather than exhaustively double-annotated. As a result, we do not report inter-annotator agreement statistics.

\subsection{Conversation History Example}
\label{appx:history_example}
Using the same running example as Appendix~\ref{appx:pipeline_example}, Table~\ref{tab:pragma-conversation-snippet} shows an excerpt from the resulting long-term conversation history.
The example spans 354 days from the first evidence session to the final query, with five relevant evidence sessions distributed over 333 days and interleaved with unrelated conversational sessions.
This illustrates how \large{\textsc{pragma}} requires models to recover and integrate temporally distributed evidence from a long, heterogeneous interaction history rather than relying on a single recent or topically concentrated context.

\subsection{Query Taxonomy Examples}
\label{appx:query_example}

Table~\ref{tab:query_examples} shows representative examples for each \large{\textsc{pragma}} query type, ranging from event-level personalization to trajectory-grounded corrective reasoning.

\section{Additional Dataset Analysis}
\subsection{Implicitness Analysis}
\label{appx:implicit}

\begin{table}[t]
\centering
\setlength{\tabcolsep}{6pt}
\resizebox{\linewidth}{!}{
\begin{tabular}{lcccc}
\toprule
\textbf{Type} & \textbf{RI} & \textbf{II} & \textbf{Score} & \textbf{n} \\
\midrule
\textit{Event-Align} & 0.6566 & 0.4120 & 0.5227 & 100 \\
\textit{Event-Correct} & 0.6131 & 0.7655 & 0.6817 & 100 \\
\textit{Trajectory-Align} & 0.6718 & 0.3470 & 0.4902 & 100 \\
\textit{Trajectory-Correct} & 0.6424 & 0.6940 & 0.6656 & 100 \\
\midrule
\textit{Overall} & 0.6460 & 0.5546 & 0.5900 & 400 \\
\bottomrule
\end{tabular}
}
\caption{{\large{\textsc{pragma}}} implicitness scores by query type.}
\label{tab:implicit_pragma}
\end{table}

\begin{table}[t]
\centering
\small
\setlength{\tabcolsep}{8pt}
\begin{tabular}{lcc}
\toprule
\textbf{Analysis} & \textbf{GPT} & \textbf{Qwen} \\
\midrule
RI vs Retrieval F1 & -0.2292 & -0.2026 \\
Implicitness vs Alignment & -0.3855 & -0.3127 \\
\bottomrule
\end{tabular}
\caption{Average Spearman correlations across methods between {\large{\textsc{pragma}}} implicitness scores and retrieval/response outcomes.}
\label{tab:implicit_corr}
\end{table}

We report an auxiliary analysis of query implicitness in \large{\textsc{pragma}}. 
The goal is to quantify how much a query depends on latent user history rather than explicitly stating the required evidence or response behavior. 
Since there are no widely used metrics for implicitness, we introduce a simple diagnostic measure to characterize the extent to which a query leaves the relevant memory evidence and intended response behavior implicit.

We decompose implicitness into two components. \textbf{Retrieval Implicitness (RI)} measures how difficult it is to recover the relevant evidence from the query surface form, using lexical and semantic overlap between the query and its supporting history.
Higher RI indicates that the needed evidence is less directly recoverable from the query alone.
\textbf{Instructional Implicitness (II)} measures whether the query explicitly signals the intended personalized reasoning behavior, such as grounding a recommendation in prior events or identifying a contradiction with a trajectory.
We use \texttt{gpt-5-mini} as a judge, providing query-type definitions and a discrete ordinal rubric to score intent explicitness. 
We convert this explicitness score into instructional implicitness (II), where higher values indicate that the intended response behavior is less explicit in the query.
We combine these components into an overall implicitness score.

Table~\ref{tab:implicit_pragma} shows that \large{\textsc{pragma}} queries are generally implicit, with an average score of 0.5900 across 400 queries. 
Corrective query types exhibit substantially higher implicitness than alignment-oriented queries: Event-Correct and Trajectory-Correct obtain scores of 0.6817 and 0.6656, respectively, compared to 0.5227 and 0.4902 for Event-Align and Trajectory-Align. 
This trend is expected, as corrective queries often appear superficially plausible unless models retrieve and reason over conflicting prior evidence.

The negative correlations in Table~\ref{tab:implicit_corr} suggest that the proposed implicitness measures are broadly aligned with the intended characteristics of \large{\textsc{pragma}} queries. 
Queries with higher retrieval implicitness tend to exhibit lower retrieval F1, while higher overall implicitness is associated with lower downstream alignment performance. 
This trend is consistent with the design goal of evaluating underspecified memory reasoning beyond explicit lexical overlap.

\subsection{Benchmark Statistics}
\label{appx:bench_stats}

Table~\ref{tab:dataset_stats} and~\ref{tab:dataset_stats_by_type} present summary statistics for {\large\textsc{pragma}}.
\large{\textsc{pragma}} contains 100 users and 400 queries, with 100 queries for each of the four query types. 
Each user has an average of 41.67 sessions, consisting of event sessions, trajectory sessions, and irrelevant
filler sessions. 
In total, the benchmark contains 4,167 sessions: 494 event sessions, 673 trajectory sessions, and 3,000 filler sessions. 
Filler sessions account for 72.0\% of all sessions, making
relevant evidence sparse within the full user history.

Each user has 4--5 event memories and 6--7 trajectory memories over two latent trajectory axes. 
Queries require 4.91 evidence items on average. 
Each query contains 44.6 tokens on average, while each user
history contains 160K tokens on average (median 160K; range 127K--194K). Token counts are computed with the \texttt{gpt-5-mini} \texttt{tiktoken} encoding over the timestamped conversation history with role prefixes. The supporting evidence spans 282.8 days on average, requiring systems to retrieve and reason over temporally distributed information.

\large{\textsc{pragma}} also covers diverse user profiles: ages range from 19 to 85, with 47 native languages, 43 citizenships, and 95 unique topics. 
These statistics reflect the benchmark's focus on long, sparse, and
heterogeneous personalized memory.

\paragraph{Benchmark scale.}
\large{\textsc{pragma}} prioritizes quality and complexity over query count.
Each of the 400 queries is grounded in a long history with temporally distributed evidence, averaging 160K tokens and 4.91 evidence items per query.
Constructing each instance requires coherence across the user profile, conversational history, distributed evidence, and personalized query.
The evidence and queries are human-validated to ensure reliable grounding and personalization across four query types and diverse user profiles.
While the pipeline can be readily scaled, human validation introduces a practical trade-off between benchmark scale and quality.

\begin{table}[t]
\centering
\setlength{\tabcolsep}{5pt}
\begin{tabular}{lc}
\toprule
\textbf{Statistic} & \textbf{Value} \\
\midrule
Users & 100 \\
Queries & 400 \\
Queries per type & 100 \\
Sessions & 4,167 \\
Sessions / user & 41.67 \\
Filler sessions & 3,000 (72.0\%) \\
Event sessions & 494 \\
Trajectory sessions & 673 \\
Events / user & 4.94 \\
Trajectory states / user & 6.73 \\
Evidence / query & 4.91 \\
Query tokens & 44.6 \\
History tokens / user & 160K \\
History tokens / user (range) & 127K--194K \\
Turns / user & 323.92 \\
Evidence span / query & 282.8 days \\
Native languages & 47 \\
Citizenships & 43 \\
Topics & 95 \\
\bottomrule
\end{tabular}
\caption{Summary statistics for {\large{\textsc{pragma}}}. Averages are reported for per-user and per-query quantities.}
\label{tab:dataset_stats}
\end{table}

\begin{table}[t]
\centering
\setlength{\tabcolsep}{5pt}
\begin{tabular}{lcc}
\toprule
\textbf{Type} & \textbf{Evidence} & \textbf{Words} \\
\midrule
Event-Align & 4.94 & 20.1 \\
Event-Correct & 3.52 & 35.6 \\
Trajectory-Align & 6.73 & 59.3 \\
Trajectory-Correct & 4.44 & 32.2 \\
\bottomrule
\end{tabular}
\caption{Average evidence count and query length by query type.}
\label{tab:dataset_stats_by_type}
\end{table}

\section{Evaluation Details}
\subsection{Evaluation Prompts}
\label{appx:eval_prompt}

All automatic evaluations are conducted using rubric-based prompts tailored to each metric and query type.
The prompts instruct the evaluator model to assess responses with respect to conversational alignment and evidence grounding while considering the provided conversational context and annotated evidence.

\vspace{10pt}
\noindent
\begin{minipage}{\linewidth}
\begin{tcolorbox}[colback=white, colframe=black!25, boxrule=0.6pt, arc=4pt, left=2pt, right=2pt, top=2pt, bottom=2pt]
\promptgroup{Evaluator System Prompt (shared)}{%
Given the query, response, and evaluation criteria, evaluate whether the response is personalized.\newline
First, briefly justify how the response satisfies each criterion.\newline
Then, assign a score for each criterion as noted in the criteria.\newline
Return the scores as an array of integers or floats, in the same order as the criteria.
}
\promptgroup{Evaluator User Prompt Template (shared)}{%
Query: \{query\}\newline
Response: \{response\}\newline
Evaluation Criteria: \{criteria\}
}
\end{tcolorbox}
\vspace{2pt}
\captionof{figure}{Shared prompts used for all evaluation calls. Top: system prompt; Bottom: user prompt template. \{criteria\} is filled with the per-query-type criteria below.}
\label{fig:eval_shared}
\end{minipage}


\noindent
\begin{minipage}{\linewidth}
\begin{tcolorbox}[colback=white, colframe=black!25, boxrule=0.6pt, arc=4pt, left=2pt, right=2pt, top=2pt, bottom=2pt]
\promptgroup{Alignment Criterion --- Event-Aligned Query}{%
Does the response recommend a next step that meaningfully builds on the user's prior attempts, rather than simply repeating past actions or suggesting something unrelated?\newline
Output 1 if yes, 0 otherwise. Only output the integer.\newline
History: \{summarized\_evidence\}
}
\promptgroup{Grounding Criterion --- Event-Aligned Query}{%
Does the response explicitly use the user's prior attempts as evidence for why this recommendation fits, rather than giving only generic advice?\newline
Output 1 if yes, 0 otherwise. Only output the integer.\newline
History: \{summarized\_evidence\}
}
\end{tcolorbox}
\vspace{2pt}
\captionof{figure}{Evaluation criteria for Event-Aligned queries. Top: alignment (primary); Bottom: grounding (auxiliary).}
\label{fig:eval_type1}
\end{minipage}


\noindent
\begin{minipage}{\linewidth}
\begin{tcolorbox}[colback=white, colframe=black!25, boxrule=0.6pt, arc=4pt, left=2pt, right=2pt, top=2pt, bottom=2pt]
\promptgroup{Alignment Criterion --- Event-Corrective Query}{%
Does the response identify the event-level inconsistency between the user's query and their prior experience?\newline
Output 1 if yes, 0 otherwise. Only output the integer.\newline
History: \{summarized\_evidence\}
}
\promptgroup{Grounding Criterion --- Event-Corrective Query}{%
How many of the following history items does the response explicitly refer to?\newline
Answer with the coverage score computed by (\# referred items) / (\# total history items).\newline
Output ONLY the final score as a float between 0 and 1, rounded to 2 decimal places.\newline
History: \{summarized\_evidence\}
}
\end{tcolorbox}
\vspace{2pt}
\captionof{figure}{Evaluation criteria for Event-Corrective queries. Top: alignment (primary); Bottom: grounding (auxiliary).}
\label{fig:eval_type2}
\end{minipage}

\noindent
\begin{minipage}{\linewidth}
\begin{tcolorbox}[colback=white, colframe=black!25, boxrule=0.6pt, arc=4pt, left=2pt, right=2pt, top=2pt, bottom=2pt]
\promptgroup{Alignment Criterion --- Trajectory-Aligned Query}{%
Does the response recommend a self-positioning strategy that is meaningfully aligned with the user's trajectory across both axes, and grounds this positioning in specific past experiences, including how those experiences should be emphasized, downplayed, or omitted?\newline
Output 1 if yes, 0 otherwise. Only output the integer.\newline
Progression: \{summarized\_evidence\}
}
\promptgroup{Grounding Criterion --- Trajectory-Aligned Query}{%
Does the response explicitly justify that recommendation using the user's prior progression across both axes, rather than relying on generic advice or surface-level similarity?\newline
Output 1 if yes, 0 otherwise. Only output the integer.\newline
Progression: \{summarized\_evidence\}
}
\end{tcolorbox}
\vspace{2pt}
\captionof{figure}{Evaluation criteria for Trajectory-Aligned queries. Top: alignment (primary); Bottom: grounding (auxiliary).}
\label{fig:eval_type3}
\end{minipage}

\noindent
\begin{minipage}{\linewidth}
\begin{tcolorbox}[colback=white, colframe=black!25, boxrule=0.6pt, arc=4pt, left=2pt, right=2pt, top=2pt, bottom=2pt]
\promptgroup{Alignment Criterion --- Trajectory-Corrective Query}{%
Does the response identify the trajectory-level inconsistency between the user's intended decision and their prior trajectory?\newline
Output 1 if yes, 0 otherwise. Only output the integer.\newline
History: \{summarized\_evidence\}
}
\promptgroup{Grounding Criterion --- Trajectory-Corrective Query}{%
How many of the following history items does the response explicitly refer to?\newline
Answer with the coverage score computed by (\# referred items) / (\# total history items).\newline
Output ONLY the final score as a float between 0 and 1, rounded to 2 decimal places.\newline
History: \{summarized\_evidence\}
}
\end{tcolorbox}
\vspace{2pt}
\captionof{figure}{Evaluation criteria for Trajectory-Corrective queries. Top: alignment (primary); Bottom: grounding (auxiliary).}
\label{fig:eval_type4}
\end{minipage}

\subsection{Judge Validation and Agreement}
\label{appx:gemini}

We additionally evaluate inter-judge agreement between our primary \texttt{gpt-5} evaluator and two independent evaluators, \texttt{gemini-3.1-pro-preview} and \texttt{claude-opus-4.6} with temperature 0.0 on a balanced audit subset. 
The subset contains 800 judged instances from the main comparison setting, corresponding to 5\% of the 16,000 evaluated instances in this setting.
It is balanced across two response models, two evaluation metrics, ten methods, and four query types, with five examples per cell. 
As shown in Table~\ref{tab:three-judge-agreement}, both independent evaluators show strong agreement with \texttt{gpt-5}.
\texttt{gemini-3.1-pro-preview} achieves 82.9\% exact agreement, Pearson $r=0.769$, and Spearman $\rho=0.759$, while \texttt{claude-opus-4.6} achieves 78.6\% exact agreement, Pearson $r=0.667$, and Spearman $\rho=0.650$ with \texttt{gpt-5}.
Across all three evaluators, three-way exact agreement reaches 72.3\%, with Krippendorff's $\alpha=0.697$. Exact agreement is consistently higher for alignment than grounding, suggesting that evidence-grounding judgments are more challenging while overall judgments remain substantially consistent across evaluator models.

\begin{table}[t]
\centering
\small
\setlength{\tabcolsep}{4pt}
\begin{tabular}{lrrrr}
\toprule

\multicolumn{5}{l}{\textbf{GPT--Gemini}} \\
Setting & $N$ & Exact & $r$ & $\rho$ \\
\cmidrule(lr){1-5}
Overall       & 800 & 82.9 & 0.769 & 0.759 \\
GPT response  & 400 & 83.5 & 0.778 & 0.770 \\
Qwen response & 400 & 82.3 & 0.753 & 0.742 \\
Alignment     & 400 & 89.0 & 0.792 & 0.792 \\
Grounding     & 400 & 76.8 & 0.705 & 0.702 \\

\midrule
\multicolumn{5}{l}{\textbf{GPT--Claude}} \\
Setting & $N$ & Exact & $r$ & $\rho$ \\
\cmidrule(lr){1-5}
Overall       & 800 & 78.6 & 0.667 & 0.650 \\
GPT response  & 400 & 77.0 & 0.652 & 0.632 \\
Qwen response & 400 & 80.3 & 0.684 & 0.673 \\
Alignment     & 400 & 81.5 & 0.670 & 0.670 \\
Grounding     & 400 & 75.8 & 0.654 & 0.659 \\

\midrule
\multicolumn{5}{l}{\textbf{Gemini--Claude}} \\
Setting & $N$ & Exact & $r$ & $\rho$ \\
\cmidrule(lr){1-5}
Overall       & 800 & 80.5 & 0.686 & 0.690 \\
GPT response  & 400 & 80.8 & 0.702 & 0.702 \\
Qwen response & 400 & 80.3 & 0.666 & 0.676 \\
Alignment     & 400 & 81.5 & 0.623 & 0.623 \\
Grounding     & 400 & 79.5 & 0.785 & 0.804 \\

\midrule
\multicolumn{5}{l}{\textbf{Three-judge agreement}} \\
Setting & $N$ & \multicolumn{2}{c}{Exact} & $\alpha$ \\
\cmidrule(lr){1-5}
Overall       & 800 & \multicolumn{2}{c}{72.3} & 0.697 \\
GPT response  & 400 & \multicolumn{2}{c}{71.8} & 0.692 \\
Qwen response & 400 & \multicolumn{2}{c}{72.8} & 0.698 \\
Alignment     & 400 & \multicolumn{2}{c}{76.0} & 0.674 \\
Grounding     & 400 & \multicolumn{2}{c}{68.5} & 0.709 \\

\bottomrule
\end{tabular}

\caption{Inter-judge agreement on the balanced 800-instance audit subset.
Exact denotes exact agreement (\%); $r$ and $\rho$ denote Pearson and
Spearman correlation, respectively. The final panel reports three-way
exact agreement and Krippendorff's $\alpha$ with interval distance.}
\label{tab:three-judge-agreement}
\end{table}

\begin{table*}[t]
\centering
\small
\setlength{\tabcolsep}{4pt}
\resizebox{\linewidth}{!}{
\begin{tabular}{lcccccccccc}
\toprule
\textbf{Method} & \multicolumn{2}{c}{\textbf{Event-Align}} & \multicolumn{2}{c}{\textbf{Event-Correct}} & \multicolumn{2}{c}{\textbf{Trajectory-Align}} & \multicolumn{2}{c}{\textbf{Trajectory-Correct}} & \multicolumn{2}{c}{\textbf{Overall}} \\
& \textbf{Align} & \textbf{Ground} & \textbf{Align} & \textbf{Ground} & \textbf{Align} & \textbf{Ground} & \textbf{Align} & \textbf{Ground} & \textbf{Align} & \textbf{Ground} \\
\midrule
\textit{No-Context} & 60.00 & 0.00 & 2.00 & 9.52 & 14.00 & 0.00 & 1.00 & 1.62 & 19.25 & 2.79 \\
\textit{Gold} & \textbf{100.00} & \textbf{100.00} & \textbf{98.00} & \textbf{100.00} & \textbf{99.00} & \textbf{99.00} & \textbf{100.00} & \textbf{100.00} & \textbf{99.25} & \textbf{99.75} \\
\bottomrule
\end{tabular}
}
\caption{Reference evaluations with gpt-5.}
\label{tab:reference}
\end{table*}

In Table~\ref{tab:reference}, we also report \texttt{gpt-5} evaluation results on no-context responses and gold reference responses provided in the benchmark metadata.
Gold responses consistently obtain near-ceiling scores, while no-context responses score substantially lower, suggesting that the evaluator reliably follows the intended rubrics and meaningfully distinguishes grounded personalized guidance from generic responses.

\section{Implementation and Baseline Details}
\label{appx:config}

Unless otherwise noted, experiments are evaluated under fixed retrieval settings and deterministic or near-deterministic decoding configurations.
We therefore report single-run results without variance estimates or error bars.

\subsection{Model and Retrieval Configurations}

\paragraph{Model Setup.}
For all dense retrieval and memory-system experiments, we use \texttt{BAAI/bge-base-en-v1.5} as the embedding model unless otherwise noted.
For response generation, we evaluate two backbone LLMs: \texttt{gpt-5-mini-2025-08-07} and \texttt{qwen3-30b-a3b-instruct-2507}. 
Unless otherwise noted, prompts use the same system instruction across
methods: the model is asked to generate a concise personalized response conditioned on the retrieved memories or conversation context.
For Qwen, we use deterministic decoding with temperature $0.0$ in all runs.

\paragraph{Static Retrieval Baselines.}
For retrieval baselines, we evaluate both turn-level and session-level variants.
Because retrieval units differ substantially in length, we set retrieval depth by granularity rather than using a single global top-$k$.
Turn-level baselines retrieve 10 turns per query, while session-level baselines retrieve 5 sessions per query.
The session-level depth approximately matches the typical number of evidence sessions associated with each query and avoids giving session-level baselines an excessively large context budget.
These retrieval depths were fixed before evaluation and were not tuned on the test set.

The turn-level BM25 baseline indexes each user turn and retrieves the top 10 turns using Okapi BM25 with default hyperparameters.
The turn-level dense baseline embeds each turn and retrieves the top 10 turns by cosine similarity. 
The window baseline follows a simple hierarchical retrieval strategy: it first retrieves the top 10 child turns using the same dense retriever, then expands each retrieved turn into a local context window of $\pm 2$ surrounding turns.

For session-level retrieval, BM25 indexes full sessions and retrieves the top 5 sessions using default BM25 hyperparameters.
The session-level dense baseline embeds full sessions and retrieves the top 5 sessions by cosine similarity. 
The session-level window baseline uses turn-level retrieval as an intermediate step: it retrieves the top 10 child turns and then expands each hit to its parent session. 
Since multiple retrieved turns can map to the same parent session, this yields fewer than 10 unique sessions in practice: 4.2 sessions per query on average.
Retrieved contexts are then passed to the response model using the same response-generation prompt.

\paragraph{Dynamic Retrieval Variants.}
We additionally implement three dynamic RAG variants inspired by prior retrieval-augmented generation methods: dynamic retrieval, query rewriting, and adaptive-$k$ retrieval.
All methods use \texttt{qwen3-30b-a3b-instruct} for generation.
We exclude \texttt{gpt-5-mini} from these experiments because dynamic retrieval requires token-level log probabilities.

The dynamic retrieval baseline is inspired by FLARE. 
It first retrieves an initial set of memories, then generates short look-ahead continuations and uses low-confidence generations to trigger additional
retrieval. 
We use deterministic decoding with temperature $0.0$, a maximum of six generation steps, 64 tokens per look-ahead sentence, and a low-confidence threshold of probability $0.8$. 
Each triggered retrieval retrieves the top 2 turn-level memories.
In practice, this yields 8.1 retrieved memories per query on average, close to the fixed top-10 budget used by the turn-level RAG baselines.
The session level variant retrieves 5.2 sessions per query on average.

The query-rewriting baseline is inspired by HyDE. 
It first generates a hypothetical answer passage for the query and uses that generated passage. 
Retrieval is then performed using the average of the original-query embedding and the hypothetical-passage embedding.  
We use temperature $0.7$ for hypothetical-passage generation, matching the open-ended generation setting used in HyDE-style retrieval, and generate the final answer deterministically with temperature $0.0$.

The adaptive-$k$ baseline is inspired by adaptive-$k$ retrieval. 
For each query, it computes similarities against all candidate turns and chooses the number of retrieved memories by applying a largest-gap heuristic to the sorted similarity scores. 
Following the Adaptive-$k$ implementation, we ignore the lower half of the score distribution when searching for the gap and include two additional items beyond the selected cutoff as a small buffer.
We cap the final retrieval depth at the same maximum budget as the turn-level RAG baselines, i.e., at most 10 retrieved turns.
For the session-level variant, we apply the same procedure with the corresponding session-level retrieval budget.

\subsection{Memory System Configurations}

We evaluate three memory-system baselines: A-MEM, Mem0 and SimpleMem.
For all memory systems, we ingest the full chronological history of each user before answering any query.
To isolate the effect of the final response generator, the memory-construction backbone is fixed to \texttt{qwen3-30b-a3b-instruct}, while the final response is generated with either the same model or \texttt{gpt-5-mini}.
Unless otherwise noted, Qwen-based generation uses deterministic decoding with temperature $0.0$.

\paragraph{A-MEM.}
For A-MEM, we use the official agentic memory implementation with \texttt{bge-base-en-v1.5} as the embedding model.
Each user--assistant chunk is added as an A-MEM note.
During ingestion, A-MEM converts each note into a structured memory containing the memory content, generated context, keywords, tags, temporal metadata, importance score, and links to related memories.
At query time, we retrieve the top 10 related A-MEM notes using its internal embedding retriever, which searches over structured note representations containing memory content, generated context, keywords, and tags.
The retrieved memory contents are then provided to the response generator.

\paragraph{Mem0.}
For Mem0, we split each user session into role-valid user--assistant chunks and add each chunk with its timestamp as metadata.
Mem0 stores memories in a Qdrant vector store with 768-dimensional vectors.
At query time, we call Mem0's search API with a per-user filter to retrieve the top 10 memories.
The retrieved memory strings are provided to the response generator with the original query.

\paragraph{SimpleMem.}
For SimpleMem, we add each dialogue turn with its speaker role, timestamp, and session identifier, then call SimpleMem's finalization step to build the user's memory store.
We follow SimpleMem's default settings for planning and parallel ingest/retrieval, but disable reflection-based additional retrieval to keep the retrieval budget controlled.
At query time, we use SimpleMem's retrieval interface and pass up to 10 retrieved memory contexts to the response generator.

\section{Additional Experiments}

\subsection{Additional Baseline Results}
\label{more_baselines}

\begin{table*}[t]
\centering
\scriptsize
\setlength{\tabcolsep}{2.5pt}
\begin{tabular}{llcccccccccc}
\toprule
\rowcolor[HTML]{E0E0E0}
\textbf{Method} &
& \multicolumn{2}{c}{\textbf{Event-Align}}
& \multicolumn{2}{c}{\textbf{Event-Correct}}
& \multicolumn{2}{c}{\textbf{Trajectory-Align}}
& \multicolumn{2}{c}{\textbf{Trajectory-Correct}}
& \multicolumn{2}{c}{\textbf{Overall}} \\
\rowcolor[HTML]{E0E0E0}
& & \textbf{Align} & \textbf{Ground}
& \textbf{Align} & \textbf{Ground}
& \textbf{Align} & \textbf{Ground}
& \textbf{Align} & \textbf{Ground}
& \textbf{Align} & \textbf{Ground} \\
\midrule
\rowcolor[HTML]{D8D8F0}
\multicolumn{12}{l}{\textit{GPT-5-mini}} \\
Synth    & & 96.00 & 55.00 & 53.00 & 55.65 & 65.00 & 11.00 & 72.00 & 32.72 & 71.50 & 38.59 \\
RAPTOR   & & 98.00 & 73.00 &  7.00 & 26.92 & 81.00 & 15.00 & 44.00 & 21.16 & 57.50 & 34.02 \\
LightMem & & 95.00 & 55.00 &  9.00 & 32.15 & 38.00 &  2.00 & 27.00 & 23.91 & 42.25 & 28.27 \\
\midrule
\rowcolor[HTML]{F0D8C8}
\multicolumn{12}{l}{\textit{Qwen3-30B-A3B-Instruct-2507}} \\
Synth    & & 89.00 & 55.00 & 14.00 & 33.57 & 55.00 & 11.00 & 55.00 & 28.54 & 53.25 & 32.03 \\
RAPTOR   & & 88.00 & 73.00 &  8.00 & 27.74 & 69.00 & 29.00 & 32.00 & 24.87 & 49.25 & 38.65 \\
LightMem & & 71.00 & 52.00 & 17.00 & 32.67 & 30.00 &  7.00 & 32.00 & 31.13 & 37.50 & 30.70 \\
\bottomrule
\end{tabular}
\caption{Results for additional baselines across generation models. Synth explicitly synthesizes retrieved evidence, RAPTOR uses hierarchical retrieval, and LightMem constructs summary-based memories. Align and Ground denote alignment and grounding scores, respectively.}
\label{tab:stronger-baselines}
\end{table*}

We additionally evaluate three strong baselines that capture complementary approaches to long-term memory: \textbf{LightMem}~\cite{ICLR2026_a05b7265}, a summary-based memory system; \textbf{RAPTOR}~\cite{ICLR2024_8a2acd17}, a hierarchical retrieval framework; and \textbf{Synth}, an evidence-synthesis pipeline inspired by MASS-RAG~\cite{xiao-etal-2026-mass}. 
We follow the original implementation of each method, with minor adaptations to fit our conversational benchmark. 
Table~\ref{tab:stronger-baselines} reports their performance across both generation models.

\textbf{RAPTOR} performs strongly on aligned queries but remains limited on corrective reasoning, suggesting that hierarchical memory organization alone is insufficient for corrective memory utilization.
\textbf{LightMem} provides a competitive summary-based baseline, with performance generally comparable to existing memory systems, further indicating that summarization alone does not resolve the utilization bottleneck. 
In contrast, \textbf{Synth} substantially improves performance on corrective queries compared with turn-level dense retrieval, which uses the same retrieval granularity. 
This result highlights the benefit of explicitly synthesizing retrieved evidence and suggests that improved evidence utilization can yield substantial gains even under similar retrieval settings. Nevertheless, no single approach consistently achieves strong alignment and grounding across all query types.

\subsection{Additional Generation Models}
\label{more_models}

To examine whether our findings generalize beyond the generation models used in the main experiments, we additionally evaluate three models from different families and scales: \texttt{gpt-oss-120b}~\cite{openai2025gptoss}, \texttt{llama-4-scout-17b-16e-instruct}~\cite{meta2025llama4scout}, and \texttt{claude-opus-4.6}. 
We evaluate oracle settings for all three models, along with representative retrieval and memory systems. 
Table~\ref{tab:more_models} reports the results.

Across all three models, providing summarized oracle evidence substantially improves performance over providing the original evidence sessions, reinforcing the importance of effective memory utilization. 
For \texttt{gpt-oss-120b}, overall alignment and grounding increase from 45.50 and 28.09 with Oracle-Session to 70.25 and 71.72 with Oracle-Summary. 
The gap is particularly large for \texttt{llama-4-scout-17b-16e}, increasing from 8.00/7.07 to 66.50/63.54. 
Even for the substantially stronger \texttt{claude-opus-4.6}, Oracle-Summary improves overall alignment and grounding from 74.00/67.54 to 90.75/92.28. 
These results indicate that access to relevant evidence alone does not guarantee effective utilization, even for stronger generation models.

Corrective reasoning also remains challenging across model families. 
Under practical retrieval and memory settings, \texttt{gpt-oss-120b} and \texttt{llama-4-scout-17b-16e} achieve low alignment on Event-Correct, while \texttt{claude-opus-4.6} performs substantially better but still lags behind its performance on aligned queries. 
Even with Oracle-Summary, Claude reaches only 65.00 alignment on Event-Correct, compared with 100.00 on both Event-Align and Trajectory-Align and 98.00 on Trajectory-Correct. 
This suggests that corrective guidance remains difficult even when the required evidence is explicitly available in a concise form.

Finally, the relative effectiveness of retrieval granularities and memory representations varies across generation models. 
While structured or summarized memories can substantially benefit some models and query types, stronger models such as \texttt{claude-opus-4.6} often perform well with session-level retrieval, which preserves more of the original conversational context. No single representation is consistently optimal across models and query types. 
Overall, these results reinforce our main conclusion that effective memory utilization remains a central challenge across model families and capacities.

\begin{table*}[t]
\centering
\scriptsize
\setlength{\tabcolsep}{1.4pt}
\begin{tabular}{llcccccccccc}
\toprule
\rowcolor[HTML]{E0E0E0}
\textbf{Method} & & \multicolumn{2}{c}{\textbf{Event-Align}} & \multicolumn{2}{c}{\textbf{Event-Correct}} & \multicolumn{2}{c}{\textbf{Trajectory-Align}} & \multicolumn{2}{c}{\textbf{Trajectory-Correct}} & \multicolumn{2}{c}{\textbf{Overall}} \\
\rowcolor[HTML]{E0E0E0}
& & \textbf{Align} & \textbf{Ground} & \textbf{Align} & \textbf{Ground} & \textbf{Align} & \textbf{Ground} & \textbf{Align} & \textbf{Ground} & \textbf{Align} & \textbf{Ground} \\
\midrule
\rowcolor[HTML]{F0D8C8}
\multicolumn{12}{l}{\textit{Llama-4-Scout-17B-16E-Instruct}} \\
\rowcolor[HTML]{FAF0E8}
\rowcolor[HTML]{FAF0E8}
\textit{Oracle Session} &  & 22.00 & 17.00 & 2.00 & 6.17 & 3.00 & 1.00 & 5.00 & 4.09 & 8.00 & 7.07 \\
\rowcolor[HTML]{FAF0E8}
\textit{Oracle Summary} &  & \underline{68.00} & \underline{70.00} & \underline{37.00} & \underline{55.25} & \underline{84.00} & \underline{78.00} & \underline{77.00} & \underline{50.90} & \underline{66.50} & \underline{63.54} \\
\midrule
\rowcolor[HTML]{FAF0E8}
\textit{Full Context} &  & 6.00 & 2.00 & 0.00 & 3.22 & 0.00 & 0.00 & 0.00 & 1.70 & 1.50 & 1.73 \\
\rowcolor[HTML]{FAF0E8}
\textit{Dense} & Session & 34.00 & 20.00 & 0.00 & 9.51 & 2.00 & 0.00 & 9.00 & 4.40 & 11.25 & 8.48 \\
\rowcolor[HTML]{FAF0E8}
\textit{} & Turn & \textbf{62.00} & 35.00 & 9.00 & 27.74 & 2.00 & 0.00 & 28.00 & \textbf{17.20} & 25.25 & 19.98 \\
\rowcolor[HTML]{FAF0E8}
\textit{BM25} & Session & 26.00 & 13.00 & 1.00 & 8.68 & 2.00 & 0.00 & 3.00 & 3.09 & 8.00 & 6.19 \\
\rowcolor[HTML]{FAF0E8}
\textit{} & Turn & 40.00 & 30.00 & 5.00 & 17.08 & 4.00 & 0.00 & 16.00 & 12.55 & 16.25 & 14.91 \\
\midrule
\rowcolor[HTML]{FAF0E8}
\textit{A-MEM} &  & 36.00 & 22.00 & 1.00 & 15.75 & 2.00 & 0.00 & 7.00 & 6.39 & 11.50 & 11.04 \\
\rowcolor[HTML]{FAF0E8}
\textit{SimpleMem} &  & 61.00 & \textbf{49.00} & \textbf{20.00} & \textbf{39.47} & \textbf{8.00} & \textbf{2.00} & \textbf{29.00} & 16.02 & \textbf{29.50} & \textbf{26.62} \\
\midrule
\rowcolor[HTML]{D8D8F0}
\multicolumn{12}{l}{\textit{GPT-OSS-120B}} \\
\rowcolor[HTML]{F0F0FA}
\rowcolor[HTML]{F0F0FA}
\textit{Oracle Session} &  & 74.00 & 42.00 & 2.00 & 18.71 & 81.00 & 26.00 & 25.00 & 25.67 & 45.50 & 28.09 \\
\rowcolor[HTML]{F0F0FA}
\textit{Oracle Summary} &  & \underline{100.00} & \underline{85.00} & \underline{9.00} & \underline{35.82} & \underline{100.00} & \underline{100.00} & \underline{72.00} & \underline{66.06} & \underline{70.25} & \underline{71.72} \\
\midrule
\rowcolor[HTML]{F0F0FA}
\textit{Dense} & Session & 78.00 & 45.00 & \textbf{3.00} & 13.68 & 69.00 & \textbf{18.00} & 23.00 & 17.02 & 43.25 & 23.43 \\
\rowcolor[HTML]{F0F0FA}
\textit{} & Turn & 88.00 & 36.00 & 1.00 & 18.78 & 52.00 & 4.00 & \textbf{28.00} & 20.22 & 42.25 & 19.75 \\
\rowcolor[HTML]{F0F0FA}
\textit{BM25} & Session & 73.00 & 32.00 & 1.00 & 18.08 & 54.00 & 9.00 & 16.00 & 15.35 & 36.00 & 18.61 \\
\rowcolor[HTML]{F0F0FA}
\textit{} & Turn & 66.00 & 30.00 & 2.00 & 16.87 & 55.00 & 9.00 & 22.00 & 19.79 & 36.25 & 18.91 \\
\midrule
\rowcolor[HTML]{F0F0FA}
\textit{A-MEM} &  & 83.00 & 44.00 & 2.00 & 18.16 & \textbf{73.00} & 14.00 & 24.00 & 17.90 & \textbf{45.50} & 23.52 \\
\rowcolor[HTML]{F0F0FA}
\textit{SimpleMem} &  & \textbf{91.00} & \textbf{60.00} & 1.00 & \textbf{19.12} & 56.00 & 12.00 & 20.00 & \textbf{21.58} & 42.00 & \textbf{28.18} \\
\midrule
\rowcolor[HTML]{D8E8D8}
\multicolumn{12}{l}{\textit{Claude-Opus-4.6}} \\
\rowcolor[HTML]{F0FAF0}
\rowcolor[HTML]{F0FAF0}
\textit{Oracle Session} &  & 83.00 & 89.00 & 65.00 & 70.27 & 56.00 & 52.00 & 92.00 & 58.88 & 74.00 & 67.54 \\
\rowcolor[HTML]{F0FAF0}
\textit{Oracle Summary} &  & \underline{100.00} & \underline{100.00} & \underline{65.00} & \underline{86.51} & \underline{100.00} & \underline{100.00} & \underline{98.00} & \underline{82.61} & \underline{90.75} & \underline{92.28} \\
\midrule
\rowcolor[HTML]{F0FAF0}
\textit{Full Context} &  & 92.00 & 94.00 & 44.00 & 60.43 & 73.00 & 75.00 & 78.00 & 57.45 & 71.75 & 71.72 \\
\rowcolor[HTML]{F0FAF0}
\textit{Dense} & Session & \textbf{100.00} & \textbf{100.00} & 55.00 & \textbf{73.27} & 85.00 & \textbf{77.00} & \textbf{89.00} & \textbf{60.99} & \textbf{82.25} & \textbf{77.81} \\
\rowcolor[HTML]{F0FAF0}
\textit{} & Turn & 97.00 & 86.00 & 45.00 & 66.24 & 76.00 & 42.00 & 85.00 & 40.02 & 75.75 & 58.56 \\
\rowcolor[HTML]{F0FAF0}
\textit{BM25} & Session & 97.00 & 94.00 & \textbf{57.00} & 72.01 & \textbf{88.00} & 68.00 & 82.00 & 51.54 & 81.00 & 71.39 \\
\rowcolor[HTML]{F0FAF0}
\textit{} & Turn & 92.00 & 84.00 & 36.00 & 57.03 & 77.00 & 36.00 & 79.00 & 38.08 & 71.00 & 53.78 \\
\midrule
\rowcolor[HTML]{F0FAF0}
\textit{A-MEM} &  & 99.00 & 97.00 & 47.00 & 70.02 & 87.00 & 65.00 & \textbf{89.00} & 52.89 & 80.50 & 71.23 \\
\rowcolor[HTML]{F0FAF0}
\textit{SimpleMem} &  & 96.00 & 94.00 & 40.00 & 63.85 & 75.00 & 33.00 & 70.00 & 38.48 & 70.25 & 57.33 \\
\bottomrule
\end{tabular}
\caption{Additional generation-model results using llama-4-scout-17b-16e-instruct, gpt-oss-120b, and claude-opus-4.6. Full-context results for gpt-oss are omitted because its context window is shorter than some {\large{\textsc{pragma}}} histories.}
\label{tab:more_models}
\end{table*}

\section{Qualitative Analysis}
\subsection{Memory Representation Examples}

Table~\ref{tab:memsys_storage_examples} shows representative stored memories from the same user across A-MEM, Mem0, and SimpleMem.
The examples illustrate the different storage formats used by each memory system.
A-MEM stores structured memory notes with metadata and contextual fields, Mem0 stores atomic natural-language memory entries, and SimpleMem stores structured atomic entries consisting of a lossless restatement plus metadata fields such as keywords, timestamp, location, persons, entities, and topic.

\subsection{Retrieval--Utilization Failure Cases}

Table~\ref{tab:rag_memsys_response_case_studies} shows representative cases where standard RAG systems successfully retrieved all annotated evidence but nevertheless failed to incorporate much of that information into the final response.
In contrast, memory systems often produced more complete responses on the same queries using fewer but more structured memories.
These examples suggest that personalized guidance depends not only on retrieval quality, but also on how retrieved information is organized and exposed to the response model.

\subsection{Grounding Failures in Follow-up Interactions}
\label{appx:follow-up}

Table~\ref{tab:followup-grounding} illustrates why alignment alone is insufficient for evaluating personalized guidance.
We construct the follow-up case from an existing event-aligned wellness query for which the initial response is judged aligned in both conditions.
The two conditions differ in grounding: the grounded condition uses an oracle-summary response with alignment=1 and grounding=1, while the poor-grounding condition uses a SimpleMem response with
alignment=1 and grounding=0.
For the follow-up turn, each condition includes its own first model response in chat history, and both conditions are then given the same follow-up query.  The grounded condition receives the oracle evidence summaries as memory context, whereas the poor-grounding condition performs fresh SimpleMem retrieval using the follow-up query.

During the follow-up interaction, this difference leads to substantially different recommendations.
The user asks whether they should replace a daily sweet bottled drink after hard morning runs with a sports hydration drink.
The grounded condition retains the user's prior low-salt meal plan, nutritionist consultation, and lower-sugar dietary choices, and therefore gives a conditional recommendation that cautions against
treating sports drinks as an unchecked default.
In contrast, the poor-grounding condition produces a more generic hydration recommendation: it endorses sports drinks for hard, sweaty runs and provides a generic sodium range, without connecting the
advice to the user's remembered low-salt nutrition plan.

This example highlights why alignment and grounding should be evaluated separately.
A response may appear generally aligned with the user's goals while still failing to preserve the specific constraints required for personalized guidance.
Such failures become particularly important in follow-up interactions, where earlier responses themselves become part of the conversational context used in future reasoning.

\section{Use of AI Assistants}
AI assistants were used during manuscript preparation for limited coding support, language editing, drafting assistance, and iterative refinement of phrasing and presentation.
All analyses, experimental decisions, interpretations, and final manuscript contents were reviewed and finalized by the authors.

\begin{table*}[t]
\centering
\footnotesize
\setlength{\tabcolsep}{5pt}
\renewcommand{\arraystretch}{1.12}
\begin{tabularx}{\textwidth}{
@{}
>{\raggedright\arraybackslash}p{3.0cm}
>{\raggedright\arraybackslash\hspace{0pt}}X
@{}
}
\toprule
\textbf{Pipeline stage} & \textbf{Concrete example} \\
\midrule
\textbf{Privasis-Zero} & Profile: \texttt{age=41}, \texttt{income=lower}, \texttt{language=Italian}, \texttt{citizenship=Italy}; seed event: publicly switched from Partito Democratico to Lega Nord because of local economic decline \ldots{} \\
\cmidrule{1-2}
\textbf{Topic} & \textit{Running a small business in Italy} \\
\cmidrule{1-2}
\textbf{Persona} & A 41-year-old Italian of modest means who publicly switched political allegiance over local economic and business concerns. \\
\cmidrule{1-2}
\textbf{Axes ($\times 2$)} & Principle continuity $\leftrightarrow$ strategic flexibility; local $\leftrightarrow$ national/global focus \\
\cmidrule{1-2}
\textbf{Events (2--5/user)} & Signed shop lease (2025-05-12) $\rightarrow$ registered business/VAT (2025-08-03) $\rightarrow$ first market sale (2025-11-20) $\rightarrow$ applied for fa\c{c}ade grant (2026-02-14) $\rightarrow$ hired assistant (2026-04-10) \\
\cmidrule{1-2}
\textbf{Filler topics (30/user)} & Sourdough baking; chess endgame studies; \ldots{}; ceramic glazing \\
\cmidrule{1-2}
\textbf{Trajectory (4--8 states)} & Local/principle-focused $\rightarrow$ pragmatic party switch $\rightarrow$ \ldots{} $\rightarrow$ broader SME/EU focus $\rightarrow$ results-based flexibility \\
\midrule
\rowcolor{black!5}
\textbf{Event-Align (type 1)} & ``What are the most effective low-cost steps \ldots{} to grow a small retail shop in Italy?'' \\
\cmidrule{1-2}
\rowcolor{black!5}
\textbf{Event-Correct (type 2)} & ``Now that the storefront fix-up funded by the town grant is complete \ldots{} should I prioritize upsizing the lease or hiring \ldots{}?'' History: the grant was \textbf{applied for}, not completed. \\
\cmidrule{1-2}
\rowcolor{black!5}
\textbf{Trajectory-Align (type 3)} & ``I'm running for a regional SME consortium board \ldots{} how should I position my adaptability and scope \ldots{}?'' \\
\cmidrule{1-2}
\rowcolor{black!5}
\textbf{Trajectory-Correct (type 4)} & ``I'm planning to fully back a national pro-business slate \ldots{} Does that sound like a good plan?'' This conflicts with the results-based, local/regional trajectory. \\
\midrule
\textbf{Event evidence sessions} & \textbf{User:} ``I \ldots{} signed a lease for a small shop \ldots{} What's next?'' \newline \textbf{Assistant:} ``Congrats \ldots{} the most useful next actions are \ldots{} legal/administrative checks and practical build-out \ldots{}'' \\
\cmidrule{1-2}
\textbf{Filler sessions} & Topic: chess endgame studies (2025-09-14). \textbf{User:} ``What are they \ldots{}?'' \textbf{Assistant:} ``Endgame studies are composed positions \ldots{}'' \\
\cmidrule{1-2}
\textbf{Trajectory evidence sessions} & \textbf{User:} ``I want to prepare how I approach local politics \ldots{}'' \newline \textbf{Assistant:} ``I can help you structure your approach \ldots{}'' \\
\cmidrule{1-2}
\textbf{Full sessions} & $5$ event $+ 7$ trajectory $+ 30$ filler $= 42$ chronologically sorted sessions; gold type-1 evidence indices: \texttt{[3, 12, 24, 30, 36]} \\
\cmidrule{1-2}
\textbf{PRAGMA} & 100 histories $\times$ 4 query types $= \textbf{400 queries}$, with timestamps, evidence indices, summaries, and evaluation rubrics \\
\bottomrule
\end{tabularx}
\caption{Running example of the {\large{\textsc{pragma}}} construction pipeline. Starting from one Privasis-Zero profile, the pipeline derives a topic, persona, behavioral axes, event and trajectory evidence, unrelated filler conversations, and four query types before assembling the timestamped full history.}
\label{tab:construction-pipeline-example}
\vspace{-12pt}
\end{table*}

\begin{table*}[t]
\centering
\caption{
Excerpt from a long-term {\large{\textsc{pragma}}} conversation history.
The history spans 354 days from the first evidence session to the query,
with five evidence sessions distributed over 333 days.
Intervening sessions and portions of conversations are omitted for brevity.
}
\label{tab:pragma-conversation-snippet}

\footnotesize
\setlength{\tabcolsep}{6pt}
\renewcommand{\arraystretch}{1.12}

\begin{tabularx}{0.95\textwidth}{
@{}
>{\raggedright\arraybackslash}p{3.1cm}
>{\raggedright\arraybackslash\hspace{0pt}}X
@{}
}
\toprule
\textbf{Session} & \textbf{Conversation excerpt} \\
\midrule

\textbf{Evidence 1} \\
2025-05-12 \\
\textit{Shop lease}
&
\textbf{User:} ``I actually just signed a lease for a small shop in the
central piazza of my town. I don't have much more than that right now.
What's next?''

\textbf{Assistant:} ``Congrats on signing the lease---that's a big step.
Since the lease is already signed, the most useful next actions are a mix
of legal/administrative checks and practical build-out and opening
preparations. \ldots{}''
\\

\multicolumn{2}{c}{
\scriptsize\textit{\ldots{} 83 days and intervening sessions omitted \ldots{}}
}
\\

\textbf{Evidence 2} \\
2025-08-03 \\
\textit{Registration and VAT}
&
\textbf{User:} ``I already completed the in-person step: I went to the
Comune office downtown and finished the business registration, and I
filed the VAT paperwork there as well.''

\textbf{Assistant:} ``Thanks---that helps a lot. Since you've completed
registration at the Comune and filed VAT paperwork, \ldots{} did they
issue a registration certificate or written confirmation at the office?''
\\

\multicolumn{2}{c}{
\scriptsize\textit{\ldots{} 42 days and intervening sessions omitted \ldots{}}
}
\\

\rowcolor{black!5}
\textbf{Filler} \\
2025-09-14 \\
\textit{Chess endgame studies}
&
\textbf{User:} ``I'd like to learn about chess endgame studies. What are
they, and how do they differ from regular endgame positions?''

\textbf{Assistant:} ``Endgame studies are composed positions---often
artistic and highly refined---created to illustrate a surprising idea,
a deep tactical or strategic motif, or a single instructive theme.
\ldots{}''
\\

\multicolumn{2}{c}{
\scriptsize\textit{\ldots{} 67 days and intervening sessions omitted \ldots{}}
}
\\

\textbf{Evidence 3} \\
2025-11-20 \\
\textit{First market sale}
&
\textbf{User:} ``I set up a stall at the weekend market and sold the first
batch of products. I felt really relieved afterwards.''

\textbf{Assistant:} ``That's a big milestone---congratulations.
Selling your first batch is an important validation and that relief is
understandable. To make the most of it, \ldots{} capture what you learned
while it's fresh. \ldots{}''
\\

\multicolumn{2}{c}{
\scriptsize\textit{\ldots{} 86 days and intervening sessions omitted \ldots{}}
}
\\

\textbf{Evidence 4} \\
2026-02-14 \\
\textit{Municipal grant}
&
\textbf{User:} ``I applied for a small municipal grant at the town hall
to repair the shop fa\c{c}ade.''

\textbf{Assistant:} ``Thanks---that clarifies the situation. With an
application already submitted to the town hall, here are practical next
steps and what to expect while you wait: \ldots{}''
\\

\multicolumn{2}{c}{
\scriptsize\textit{\ldots{} 55 days and intervening sessions omitted \ldots{}}
}
\\

\textbf{Evidence 5} \\
2026-04-10 \\
\textit{Staffing and foot traffic}
&
\textbf{User:} ``I recently hired a part-time assistant---and since then
I've noticed increased foot traffic on Via Roma near the shop. \ldots{}''

\textbf{Assistant:} ``That's a useful development to work with. Since
you've observed increased foot traffic near your shop on Via Roma after
hiring the part-time assistant, here are practical, no-cost or low-cost
steps you can take to convert more of that passersby traffic into sales
while keeping labor costs controlled. \ldots{}''
\\

\multicolumn{2}{c}{
\scriptsize\textit{\ldots{} 21 days and intervening sessions omitted \ldots{}}
}
\\

\midrule
\rowcolor{black!8}
\textbf{Query} \\
2026-05-01 \\
\textit{Low-cost shop growth}
&
\textbf{User:} ``What are the most effective low-cost steps I should take
over the next couple of months to grow a small retail shop in Italy?''
\\

\bottomrule
\end{tabularx}
\end{table*}

\begin{table*}[t]
\centering
\small
\setlength{\tabcolsep}{5pt}
\renewcommand{\arraystretch}{1.15}
\begin{tabular}{p{0.20\linewidth}p{0.38\linewidth}p{0.36\linewidth}}
\toprule
\textbf{Query Type} & \textbf{User Query} & \textbf{Personalized Response (truncated)} \\
\midrule
\textbf{Event-Aligned Query} & What would you recommend I try next in seasonal Finnish baking or home cooking as early summer begins? & Since you have baked with your own rhubarb, foraged blueberries, and tackled Karelian pies, try a savory nokkospiirakka next. It builds on your pie skills and leans into early-summer foraging... \\
\midrule
\textbf{Event-Corrective Query} & Didn't we make stuffed flatbreads one night with mustard oil? Any recommendations for similar late-night bites? & I think you are mixing a couple of meals: you made aloo paratha in the morning, the mustard oil was with Bengali macher jhol at lunch, and the late-night first try was litti chokha. For similar bites... \\
\midrule
\textbf{Trajectory-Aligned Query} & I am drafting a short investor profile for regional brokerage onboarding. How should I position my control mode and geographic scope given my investing history? & Position yourself as a hands-on, Jordan-focused investor with a strict six-month cash buffer who uses selective, rules-based automation and modest regional diversification. Highlight your manual allocations, sukuk fund, and 5\% robo transfer... \\
\midrule
\textbf{Trajectory-Corrective Query} & I am planning to personally host a weekly public Q\&A at the community center every Saturday to keep everyone informed and coordinated. What do you think? & Promising intent, but a fixed weekly Q\&A conflicts with the playbook you have built: shifting to phone trees during water cuts, letting younger spokespeople lead forums, and moving from kiosks to discreet home meetings... \\
\bottomrule
\end{tabular}
\caption{Examples and truncated gold responses for each {\large{\textsc{pragma}}} query type.}
\label{tab:query_examples}
\end{table*}

\begin{table*}[t]
\centering
\scriptsize
\setlength{\tabcolsep}{4pt}
\renewcommand{\arraystretch}{1.12}
\begin{tabular}{p{0.16\linewidth}p{0.78\linewidth}}
\toprule
\textbf{System} & \textbf{Example stored memory from the same user} \\
\midrule
A-MEM
& \textbf{content:} ``User: I actually just signed a lease for a small shop in the central piazza of my town. I don't have much more than that right now. What's next? Assistant: Congrats on signing the lease ...'' \newline
\textbf{context:} the lease-signing turn is linked to shop-opening planning, permits, budget/cashflow, fit-out, insurance, suppliers, staffing, and launch marketing. \newline
\textbf{keywords/tags:} retail space, business setup, opening timeline, permits, budget, marketing strategy. \newline
\textbf{timestamp:} 2025-05-12; \textbf{importance:} 1.0. \\
\midrule
Mem0
& ``User is planning to open a small retail shop in a central piazza and was provided a concise marketing checklist with low-cost, high-impact tactics for pre-opening, soft opening, grand opening, and ongoing promotion.'' \\
\midrule
SimpleMem
& \textbf{lossless\_restatement:} ``A user initiated a conversation on 2025-05-12 to seek guidance on opening a small retail space after signing a lease for a shop in the central piazza of their town.'' \newline
\textbf{keywords:} small retail space, signed lease, central piazza, opening guidance. \newline
\textbf{timestamp/topic:} 2025-05-12; retail shop opening guidance. \\
\bottomrule
\end{tabular}
\caption{Examples of how memory systems store information for the same user.}
\label{tab:memsys_storage_examples}
\end{table*}

\begin{table*}[t]
\centering
\scriptsize
\setlength{\tabcolsep}{4pt}
\renewcommand{\arraystretch}{1.12}
\begin{tabular}{p{0.085\linewidth}p{0.185\linewidth}p{0.20\linewidth}p{0.07\linewidth}p{0.155\linewidth}p{0.20\linewidth}p{0.065\linewidth}}
\toprule
\textbf{Type} & \textbf{User query} & \textbf{Failed RAG response} & \textbf{RAG retr.} & \textbf{Why RAG fails} & \textbf{Aligned memory response} & \textbf{Mem. retr.} \\
\midrule
Event-Align
& I'm planning another short local trip in Greece---what 1--2 day destinations would you recommend that offer good walks and historic sites?
& \textit{Window}: Great---two strong 1--2-day picks: Nafplio + Epidaurus \ldots{} Delphi \ldots{} Want a sample itinerary from Athens or another starting point?
& 5/5
& Recommends Nafplio/Epidaurus and Delphi, which the user already visited, instead of proposing a next step that builds on those trips.
& \textit{A-MEM}: Nice---since you like good walks, sunsets and historic sites, try \ldots{} Meteora \ldots{} Hydra \ldots{} Monemvasia \ldots{}
& 4/5 \\
\midrule
Event-Correct
& When I was in Belgium, didn’t we digitize parish marriage books for Sicilian miners’ families? Which databases or finding aids should I use now to verify and expand that work?
& \textit{Dense}: If you digitized them in Charleroi, the images/metadata are most likely held by the local deposit of the Belgian State Archives \ldots{} Also search FamilySearch’s Belgium collections \ldots{}
& 3/3
& Accepts the false premise. The history says the Belgium work was cataloging Abruzzese miners' letters, not digitizing Sicilian parish marriage books.
& \textit{A-MEM}: Do you mean those parish marriage books were digitized while you were in Charleroi, or are you remembering only the letters? \ldots{}
& 2/3 \\
\midrule
Trajectory-Align
& I’m applying to be a mindfulness facilitator at a community center. For my short bio, how should I present where I sit on learning autonomy and geographic scope so it comes across clearly, and what should I emphasize, downplay, or omit?
& \textit{BM25}: Lead with a clear, consistent frame: ``community-anchored facilitator who cultivates learner autonomy'' \ldots{} emphasize self-directed pathways, local relationships, and remote resources \ldots{}
& 6/7
& Gives a generic trajectory label but does not reflect the user's specific progression: Amharic dawn practice, grief journaling, local radio clips, community circles, and mentoring.
& \textit{A-MEM}: Suggested one-line bio: ``I teach self-directed, Amharic-language mindfulness rooted in dawn neighborhood walks and journaling'' \ldots{} emphasize autonomy, local cultural fit, and trauma-informed skills \ldots{}
& 4/7 \\
\midrule
Trajectory-Correct
& I’m planning to accept a year-long perioperative anesthesia research fellowship in the U.S. starting this fall to deepen my specialization. What do you think?
& \textit{Dense}: Sounds like a promising step \ldots{} confirm clinical exposure, a named mentor, and funding/visa \ldots{} If those three criteria are met, I’d recommend accepting.
& 4/4
& Does not surface the trajectory conflict: the user had been emphasizing rootedness in Krak\'ow, family caregiving, and a generalist internal-medicine path.
& \textit{A-MEM}: This can be high-yield \ldots{} but consider three core trade-offs first: time away from Krak\'ow (family caregiving), whether the fellowship is research vs clinical, and funding/visa logistics \ldots{}
& 2/4 \\
\bottomrule
\end{tabular}
\caption{gpt-5-mini response cases by query type. The RAG baseline retrieved all or nearly all gold evidence but still failed the alignment judgment. The memory-system response on the same query is judged alignment-correct.}
\label{tab:rag_memsys_response_case_studies}
\end{table*}

\begin{table*}[t]
\centering
\small
\setlength{\tabcolsep}{4pt}
\renewcommand{\arraystretch}{1.15}
\begin{tabularx}{\textwidth}{p{0.16\textwidth}p{0.39\textwidth}p{0.39\textwidth}}
\toprule
\textbf{Item} & \textbf{Grounded condition} & \textbf{Poor-grounding condition} \\
\midrule
Source case
&
\multicolumn{2}{p{0.80\textwidth}}{
\textbf{User query:} ``For healthy aging and daily wellness, what would you recommend I add or adjust in my routine?''
} \\
\midrule
Relevant prior evidence
&
\multicolumn{2}{p{0.80\textwidth}}{
The user had started 30-minute dawn walks at Saras Baug in Pune, carried a water bottle and used hydration reminders, met a geriatric nutritionist to tune a fiber-rich \emph{low-salt} meal plan, received a flu shot, and chose lighter \emph{low-sugar} Diwali snacks.
} \\
\midrule
First response used as chat history
&
\textbf{Oracle-summary, GPT-5-mini; alignment=1, grounding=1.}
\emph{``Nice routine so far -- those dawn walks, hydration reminders and the nutrition plan are great foundations. To boost healthy aging, add: \ldots Keep hydration year-round (adjust electrolytes in hot weather) \ldots Stay up to date on vaccines (you already had flu \ldots), \ldots group walks or classes at Saras Baug.''}
&
\textbf{SimpleMem, GPT-5-mini; alignment=1, grounding=0.}
\emph{``Keep the basics--consistent sleep (7--8 hrs), daily moderate activity (30 min most days), twice-weekly strength work \ldots Add cognitive and fine-motor practices you already enjoy (your Raga Yaman routine and short daily drawing/pen drills) \ldots Follow a mostly Mediterranean-style diet \ldots''}
\\
\midrule
Follow-up query
&
\multicolumn{2}{p{0.80\textwidth}}{
``I've been drinking a sweet bottled drink every day after hard morning runs, so I'm planning to switch to a sports hydration drink as my default. Does that sound better?''
} \\
\midrule
Expected grounded behavior
&
\multicolumn{2}{p{0.80\textwidth}}{
Acknowledge that a sports hydration drink may be reasonable after long or very hard runs, but do not endorse it as an unchecked default. The answer should preserve the user's lower-sugar habit and nutritionist-guided low-salt constraint, check sugar/sodium labels, and recommend consulting the clinician or nutritionist if this becomes daily.
} \\
\midrule
Observed follow-up response
&
\textbf{Preserves the grounded constraint.}
\emph{``Short answer: maybe -- sports drinks can be better than a sugary bottled drink for replacing electrolytes after long, hard runs, but not always. \ldots Watch sugar \ldots Watch sodium: because you follow a low-salt meal plan (and you've been working with a geriatric nutritionist), check the sodium on the label. \ldots If you have hypertension, kidney disease, or are on a salt-restricted plan, run this change by your GP/nutritionist \ldots''}
&
\textbf{Generic endorsement.}
\emph{``Yes -- switching to a proper sports/hydration drink can be better after hard, sweaty morning runs \ldots for long or very sweaty sessions a sports drink with \textasciitilde{}4--8\% carbs (40--80 g/L) and 200--600 mg sodium/L helps rehydrate and replace electrolytes \ldots''}
\\
\midrule
Interpretation
&
Grounding turns the follow-up into a conditional recommendation constrained by remembered user-specific evidence.
&
Alignment alone is insufficient: the answer is reasonable-sounding but loses the personalized constraint that should change the recommendation.
\\
\bottomrule
\end{tabularx}
\caption{
Case study of grounding failure in follow-up interaction. Both conditions include the corresponding first model response in chat history. For the follow-up turn, the grounded condition uses oracle
evidence summaries as memory context, whereas the poor-grounding condition uses fresh SimpleMem retrieval on the follow-up query.
}
\label{tab:followup-grounding}
\end{table*}

\end{document}